%% file: main.tex
\documentclass[journal]{IEEEtran} 

\usepackage{amsmath,amssymb,amsfonts,booktabs,multirow,bm,balance,bbm,siunitx,hyperref,graphicx,algorithm,makecell}
\usepackage[noend]{algpseudocode}
\usepackage[caption=false, font=footnotesize]{subfig}
\usepackage[usenames,dvipsnames]{color}

\input{macros}

\begin{document}

\title{Design and Control of a Midair-Reconfigurable Quadcopter using Unactuated Hinges}

\author{Nathan~Bucki,~\IEEEmembership{Student~Member,~IEEE,}
Jerry~Tang,~\IEEEmembership{Student~Member,~IEEE,}
and~Mark~W.~Mueller,~\IEEEmembership{Member,~IEEE}
\thanks{The authors are with the High Performance Robotics Lab at the University of California, Berkeley, Berkeley,
	CA, 94709 e-mail: {\tt\small \{nathan\_bucki, jerrytang, mwm\}@berkeley.edu}.}
}

\maketitle
\thispagestyle{empty}
\pagestyle{empty}

\begin{abstract}
A novel quadcopter capable of changing shape mid-flight is presented, allowing for operation in four configurations with the capability of sustained hover in three.
This is accomplished without requiring actuators beyond the four motors typical of a quadcopter.
Morphing is achieved through freely-rotating hinges that allow the vehicle arms to fold downwards by either reducing or reversing thrust forces.
Constraints placed on the control inputs of the vehicle prevent the arms from folding or unfolding unexpectedly.
This allows for the use of existing quadcopter controllers and trajectory generation algorithms with only minimal added complexity.
For our experimental vehicle at hover, we find that these constraints result in a 36\% reduction of the maximum yaw torque the vehicle can produce, but do not result in a reduction of the maximum thrust or roll and pitch torques.
Experimental results show that, for a typical maneuver, the added limits have a negligible effect on trajectory tracking performance.
Finally, the ability to change configurations is shown to enable the vehicle to traverse small passages, perch on hanging wires, and perform limited grasping tasks.
\end{abstract}

\input{introduction}
\input{sysModel}
\input{control}

\input{design}
\input{experiment}
\input{conclusion}

\section*{Acknowledgement}
This material is based upon work supported by the National Science Foundation Graduate Research Fellowship under Grant No. DGE 1752814 and by the Berkeley Fellowship for Graduate Study.
The experimental testbed at the HiPeRLab is the result of contributions of many people, a full list of which can be found at \url{hiperlab.berkeley.edu/members/}.

\bibliographystyle{IEEEtran}
\bibliography{references}

\end{document}

%% file: macros.tex
\newcommand{\bs}{\boldsymbol}

\newcommand{\roundB}[1]{\left( #1 \right)} 
\newcommand{\curlyB}[1]{\left\{ #1 \right\}} 
\newcommand{\absB}[1]{\left| #1 \right|} 
\newcommand{\skewSymm}[1]{\bs S\!\roundB{#1}} 

\newcommand{\z}[2]{\ensuremath{\bs z_{#1}^{#2}}}
\newcommand{\y}[2]{\ensuremath{\bs y_{#1}^{#2}}}
\newcommand{\x}[2]{\ensuremath{\bs x_{#1}^{#2}}}
\newcommand{\force}[2]{\ensuremath{\bs f_{#1}^{#2}}}
\newcommand{\torque}[2]{\ensuremath{\bs \tau_{#1}^{#2}}}
\newcommand{\thrust}[1]{\ensuremath{f_{p_{#1}}}}
\newcommand{\propTorque}[1]{\ensuremath{\tau_{p_{#1}}}}
\newcommand{\rot}[1]{\ensuremath{\bs R^{#1}}}
\newcommand{\dist}[2]{\ensuremath{\bs d_{#1}^{#2}}}

\newcommand{\distddot}[2]{\ensuremath{\bs{\ddot{d}}_{#1}^{#2}}}

\newcommand{\distx}[2]{\ensuremath{d_{#1,x}^{#2}}}
\newcommand{\disty}[2]{\ensuremath{d_{#1,y}^{#2}}}
\newcommand{\distz}[2]{\ensuremath{d_{#1,z}^{#2}}}
\newcommand{\angvel}[2]{\ensuremath{\bs \omega_{#1}^{#2}}}
\newcommand{\angacc}[2]{\ensuremath{\bs{\dot{\omega}}_{#1}^{#2}}}
\newcommand{\inertia}[2]{\ensuremath{\bs J_{#1}^{#2}}}
\newcommand{\inertiaxx}[2]{\ensuremath{J_{#1,xx}^{#2}}}
\newcommand{\inertiayy}[2]{\ensuremath{J_{#1,yy}^{#2}}}
\newcommand{\inertiazz}[2]{\ensuremath{J_{#1,zz}^{#2}}}
\newcommand{\mass}[1]{\ensuremath{m_{#1}}}
\newcommand{\armAngle}{\ensuremath{\theta}}
\newcommand{\thrustToTorqueForward}{\ensuremath{\kappa^{+}}}
\newcommand{\thrustToTorqueBack}{\ensuremath{\kappa^{-}}}
\newcommand{\thrustToTorque}[1]{\ensuremath{\kappa_{p_{#1}}}}
\newcommand{\totalThrust}{\ensuremath{f_\Sigma}}
\newcommand{\bodyTorque}{\ensuremath{\bs \tau^B}}
\newcommand{\thrusts}{\ensuremath{u}}
\newcommand{\attError}{\ensuremath{\bs r}}
\newcommand{\stateCost}{\ensuremath{Q}}
\newcommand{\rollPitchCost}{\ensuremath{q_{\phi_e \theta_e}}}
\newcommand{\yawCost}{\ensuremath{q_{\psi_e}}}
\newcommand{\rollPitchRateCost}{\ensuremath{q_{\dot{\phi}_e \dot{\theta}_e}}}
\newcommand{\yawRateCost}{\ensuremath{q_{\dot{\psi}_e}}}
\newcommand{\inputCost}{\ensuremath{R_{\bodyTorque}}}
\newcommand{\forwardThrustCost}{\ensuremath{r^{+}}}
\newcommand{\reverseThrustCost}{\ensuremath{r^{-}}}
\newcommand{\inputCostThrusts}{\ensuremath{R_{\thrusts}}}
\newcommand{\unfoldedMixer}{\ensuremath{M_{u}}}
\newcommand{\twoArmsMixer}{\ensuremath{M_{2f}}}
\newcommand{\fourArmsMixer}{\ensuremath{M_{4f}}}
\newcommand{\minVehicleDim}{\ensuremath{d}}
\newcommand{\ineqMatrix}{\ensuremath{W}}
\newcommand{\ineqMatrixUnfolded}{\ensuremath{W_u}}
\newcommand{\ineqMatrixTwoArms}{\ensuremath{W_{2f}}}
\newcommand{\minThrust}{\ensuremath{f_{\textrm{min}}}}
\newcommand{\maxThrust}{\ensuremath{f_{\textrm{max}}}}
\newcommand{\desThrust}{\ensuremath{f_{\textrm{des}}}}
\newcommand{\totalInertia}{\inertia{\Sigma}{B}}
\newcommand{\totalInertiaxx}{\inertiaxx{\Sigma}{B}}
\newcommand{\totalInertiayy}{\inertiayy{\Sigma}{B}}
\newcommand{\totalInertiazz}{\inertiazz{\Sigma}{B}}

%% file: introduction.tex
\section{Introduction} \label{sec:intro}
\IEEEPARstart{I}{n} recent years, quadcopters have proven to be useful in performing a number of tasks such as building inspection, surveillance, package delivery, and search and rescue.
Many extensions of the original quadcopter design have been proposed in order to allow for new tasks to be performed, improving their utility.
However, this typically requires the vehicle to carry additional hardware, which not only can reduce flight time due to the increased weight of the system, but can also increase the complexity of the vehicle, making it more difficult to build and maintain, which can lead to a higher likelihood of system failures.
In this work we present a design change to the quadcopter which allows the vehicle to change shape during flight, perch, and perform simple aerial manipulation, all without requiring significant hardware additions (e.g. motors or complex mechanisms).

\subsection{Related Work}\label{sec:relWork}

Several aerial vehicles capable of changing shape have been previously developed.
For example, in \cite{bouman2019design} a vehicle capable of automatically unfolding after being launched from tube is presented, and in \cite{mintchev2015foldable} a vehicle is presented which uses foldable origami-style arms to automatically increase its wingspan during takeoff. 
Although such designs excel in enabling the rapid deployment of aerial vehicles, they do not focus on repeated changes of shape, and thus require intervention to be returned to their compressed forms.

Vehicles capable of changing shape mid-flight have also been developed in order to enable the traversal of narrow passages.
In \cite{zhao2017deformable} a vehicle that uses several servomotors to actuate a scissor-like structure that can shrink or expand the size of the vehicle is presented, and in \cite{yang2019design} a single servomotor is used in conjunction with an origami structure to enable the arms of a quadcopter to shorten or lengthen during flight.
Vehicles that use a central actuator to change the angle of their arms in an X-shape are presented in \cite{desbiez2017x} and \cite{bai2019evaluation}, and a vehicle that uses four servomotors to change each arm angle is presented in \cite{falanga2018foldable} and extended in \cite{fabris2020geometry}.
In \cite{vargas2015dynamic} and \cite{sakaguchi2019novel} a quadcopter design is presented that is capable of using one or more actuators to reposition the propellers of the vehicle to be above one another such that the horizontal dimension of the vehicle is reduced.
Similarly, \cite{riviere2018agile} uses a single actuator to reposition the propellers of the vehicle to be in a horizontal line, and demonstrates the vehicle being used to traverse a narrow gap.

Several designs have also been proposed that enable aerial vehicles to perch on structures in the environment.
Such vehicles are able to fly to a desired location, attach themselves to a feature in the environment, and then remain stationary without consuming significant amounts of energy (e.g. while monitoring the surrounding area).
In \cite{hawkes2013dynamic} a passive adhesive mechanism is proposed for perching on smooth surfaces, and in \cite{kalantari2015autonomous} adhesive pads are used in conjunction with a servomotor to attach and detach the vehicle from vertical walls.
In \cite{popek2018autonomous} and \cite{hang2019perching} grippers actuated using servomotors are used to enable perching on bars.
Similarly, \cite{doyle2012avian} describes a purely passive gripper that used the weight of the vehicle to close a gripper around a horizontal bar.

Finally, a large amount of work has been produced regarding the use of quadcopters to perform aerial manipulation.
Aerial vehicles with the capability to interact with the environment open the door to a wide range of potential applications, e.g. performing construction as shown in \cite{lindsey2012construction}.
Typically such designs involve attaching one or more robot arms to a quadcopter, as shown in \cite{thomas2013avian}, \cite{orsag2013stability}, and \cite{korpela2014towards} for example.
Similar to the design presented in this work, several existing designs involve changing the structure of the vehicle in order grasp objects, such as \cite{falanga2018foldable}, \cite{zhao2018deformable}, and \cite{anzai2018aerial}.
In \cite{falanga2018foldable} four servomotors are used to wrap the arms of the vehicle around and object in order to lift it.
Similarly, in \cite{zhao2018deformable} a novel deformable quadcopter is presented which uses a scissor-mechanism to expand and contract the body of the vehicle, allowing for both grasping and enhanced agility/stability by changing shape mid-flight.
In \cite{anzai2018aerial} a ring-shaped multicopter-like vehicle is presented which is capable of changing the overall shape of the vehicle to allow for large objects to be lifted using multiple grippers.
Other designs, such as \cite{mellinger2011design}, use passive elements to engage a gripper and a single actuator to disengage the gripper.
However, all such designs require the vehicle to carry one or more actuators beyond the four motors used to drive the propellers (e.g. servomotors used for opening/closing a gripper), increasing the weight of the vehicle and therefore decreasing flight time.
Additional examples of vehicles used to perform aerial manipulation can be found in the aerial manipulation survey papers \cite{mengsurvey} and \cite{khamseh2018aerial}.

\subsection{Capabilities of the novel vehicle}

\begin{figure}
	\centering
	
	\subfloat[]{\label{subfig:unfolded}\includegraphics[width=0.49\columnwidth]{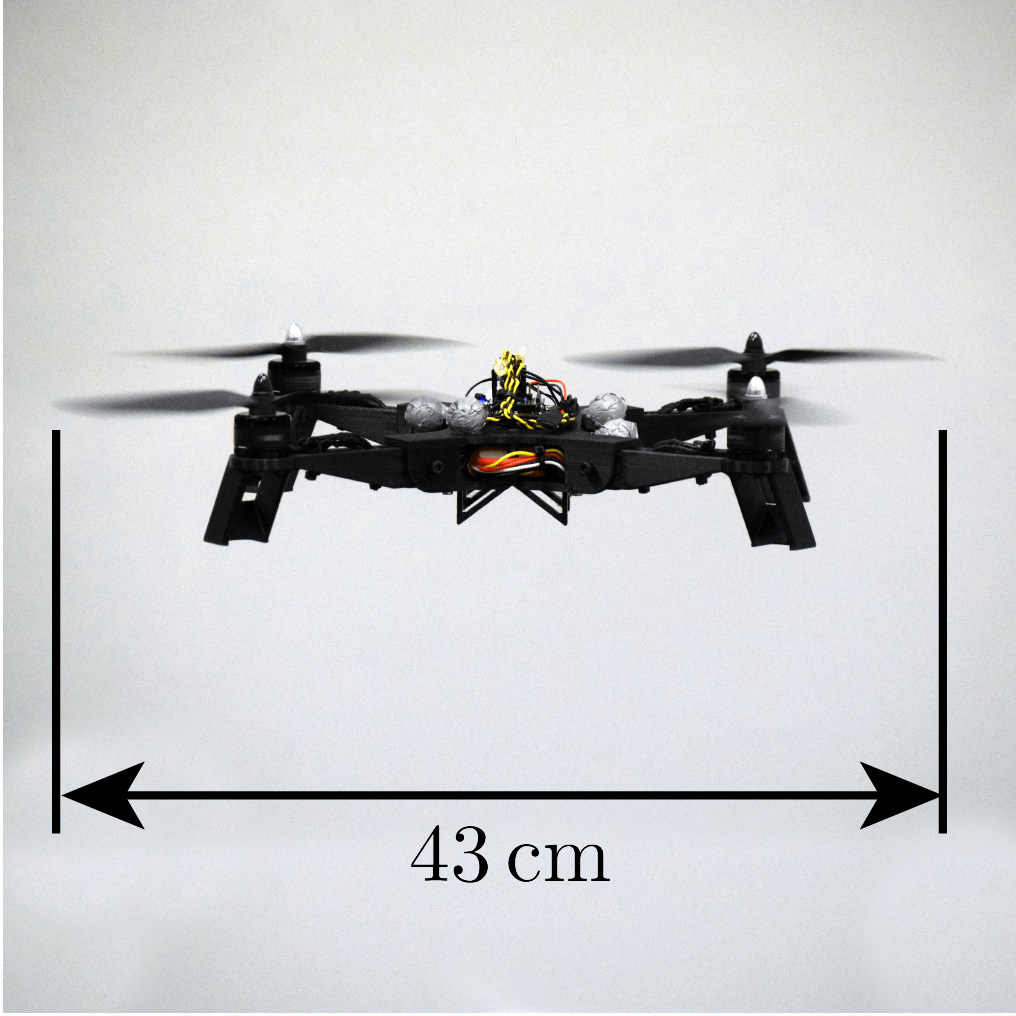}}
	\hfill
	\subfloat[]{\label{subfig:tunnel}\includegraphics[width=0.49\columnwidth]{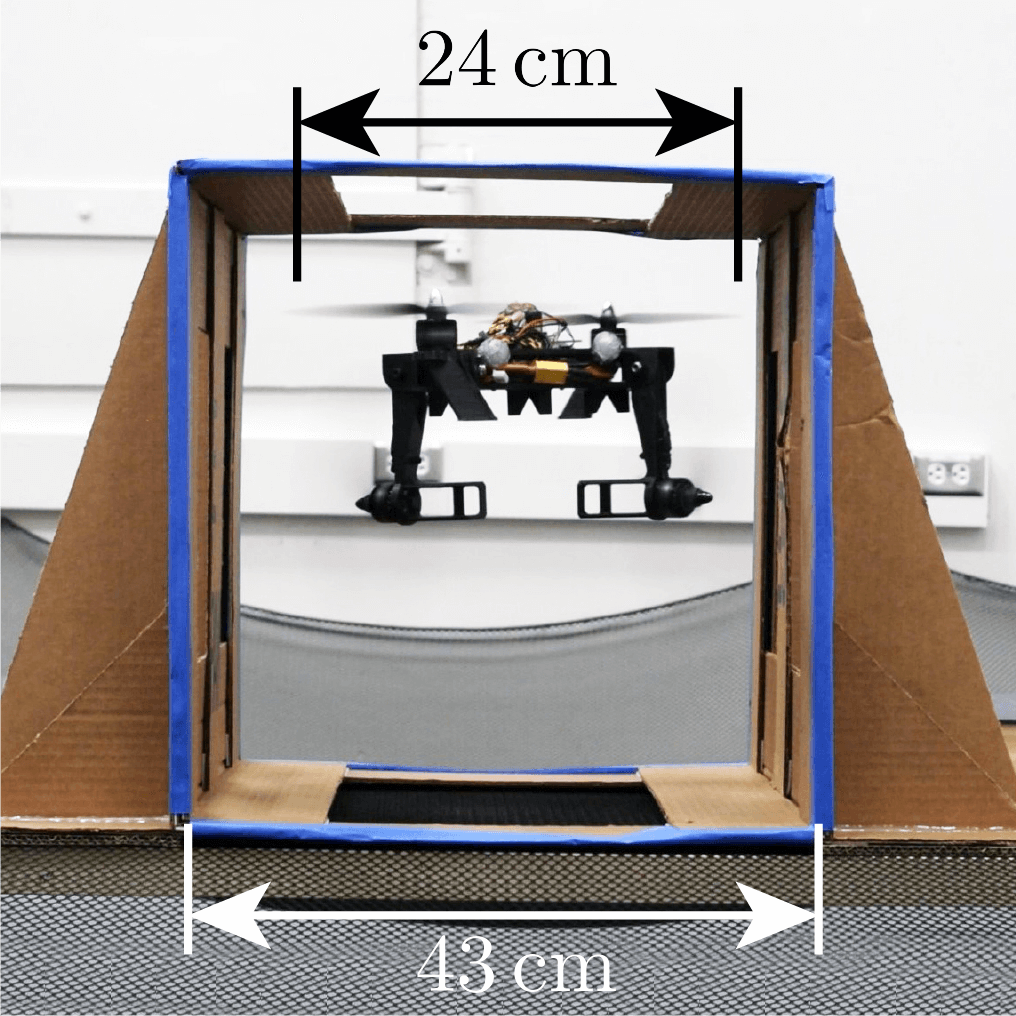}}
	\hfill
	\subfloat[]{\label{subfig:projectile}\includegraphics[width=\columnwidth]{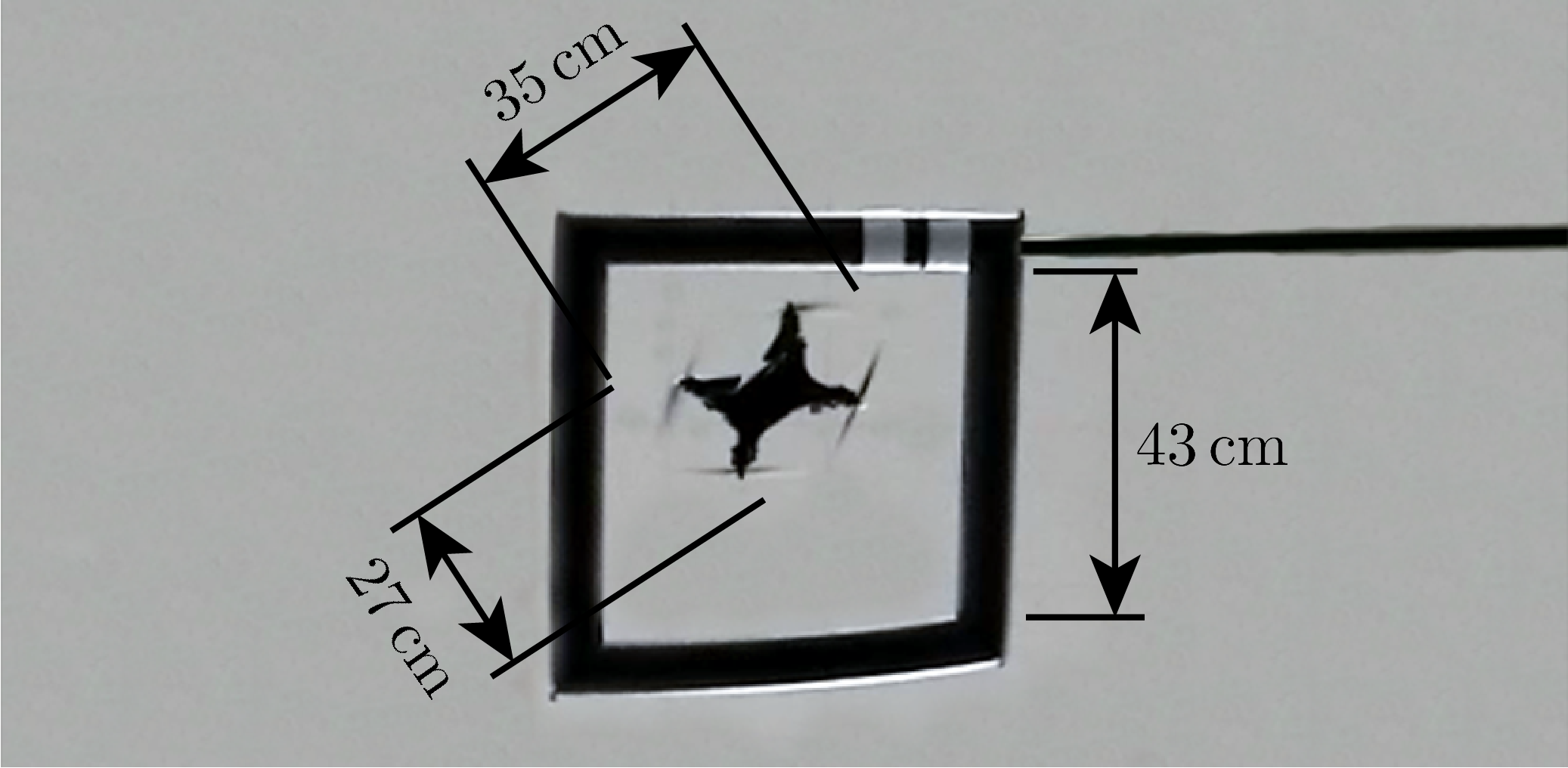}}
	\hfill
	\subfloat[]{\label{subfig:perch}\includegraphics[width=0.49\columnwidth]{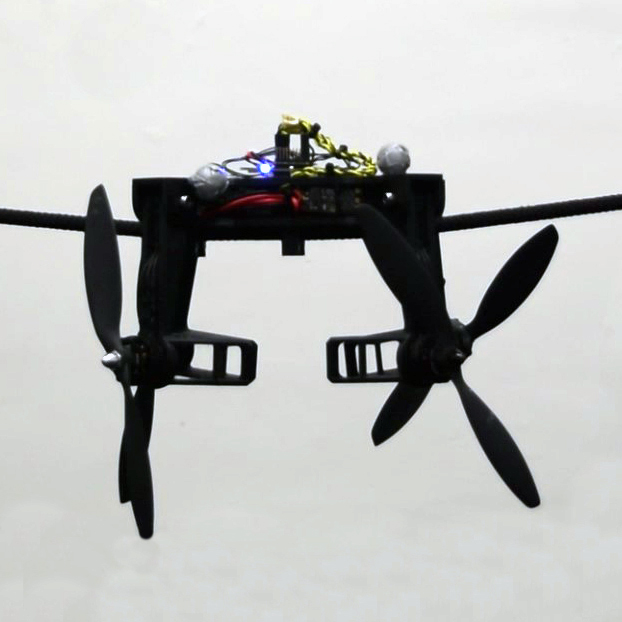}}
	\hfill
	\subfloat[]{\label{subfig:box}\includegraphics[width=0.49\columnwidth]{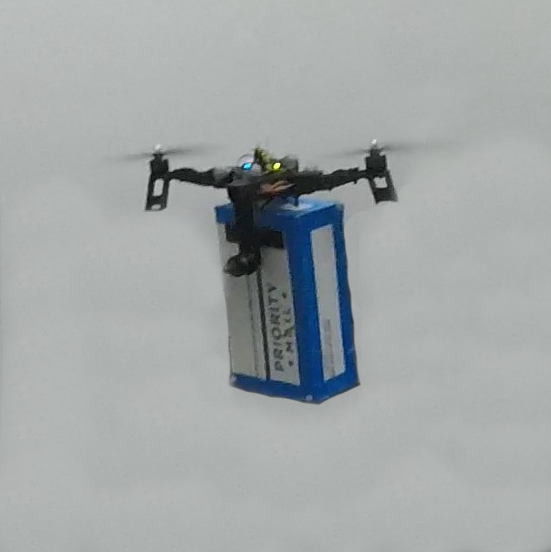}}

	\caption{Images of the experimental vehicle performing a variety of different tasks. The vehicle is capable of flying like a conventional quadcopter when in the unfolded configuration \protect\subref{subfig:unfolded}, but when flying in the two-arms-folded configuration is able to, e.g., traverse narrow tunnels \protect\subref{subfig:tunnel} and perform simple aerial manipulation tasks such as carrying a box \protect\subref{subfig:box}. Additionally, by allowing all four arms to fold, the vehicle is able to perch on thin wires \protect\subref{subfig:perch}, and even traverse narrow gaps in projectile motion \protect\subref{subfig:projectile} (view from below).}
	\label{fig:vehiclePic}
\end{figure}

In this work, we extend our prior work \cite{bucki2019design} in several ways, enabling the vehicle to perform several of the previously mentioned tasks while requiring only minor changes to the design and control of the vehicle compared to a conventional quadcopter.

In \cite{bucki2019design} a quadcopter design was presented that replaced the typically rigid connections between the arms of the quadcopter and the central body with sprung hinges that allow for the arms of the quadcopter to fold downward when low thrusts are produced by the propellers.
This feature enabled the vehicle to reduce its largest dimension while in projectile motion, allowing the vehicle to fly towards a narrow gap, collapse its arms, and then unfold after traversing the gap.
In this work we perform two significant design changes.
First, we remove the springs used to fold the arms, and instead fold each arm by reversing the thrust direction of the attached propeller.
Second, we change the geometry of the vehicle such that when two opposite arms are folded, the thrust vectors of the associated propellers are offset from one another, allowing for a yaw torque to be produced when the thrust direction of the propellers is reversed.
This contrasts works such as \cite{mueller2016relaxed} and \cite{lippiello2014emergency} which demonstrate the control of quadcopters using only two propellers to produce thrust at the expense of leaving the yaw torque unbalanced, and therefore the vehicle yaw angle uncontrolled.

By removing the springs used to fold the propellers, the proposed vehicle is able to produce a higher yaw torque and a lower minimum total thrust than the vehicle presented in \cite{bucki2019design} when flying in the unfolded configuration (i.e. as a conventional quadcopter).
For example, while the experimental vehicle presented in \cite{bucki2019design} had a maximum yaw torque reduction of 77\% at hover compared to a conventional quadcopter of similar size, the experimental vehicle presented in this work has only a 36\% reduction.
Similarly, in order to prevent its arms from folding due to the spring forces, the vehicle presented in \cite{bucki2019design} was required to produce a minimum total thrust of 70\% of the thrust produced at hover.
However, the vehicle presented in this work has no such requirement due to the lack of spring forces acting on its arms, and (similar to a conventional quadcopter) is capable of producing a minimum total thrust of zero.
Furthermore, by removing the springs, the design and manufacturing complexity of the vehicle is reduced.
 
No actuators or mechanisms other than the four hinges are added to the proposed vehicle, keeping its mass low, and only standard off-the-shelf components (e.g. propellers, motors, and electronic speed controllers) are used in the design, with the exception of the custom frame of the vehicle.
Thus, the main difference between the vehicle described in this paper and a conventional quadcopter is the fact that each arm is attached to the central body via a rotational joint rather than with a rigid connection.

These changes enable the vehicle to perform a number of tasks as shown in Figure \ref{fig:vehiclePic}, namely:
\begin{enumerate}
	\item Sustained flight as a conventional quadcopter with all four arms unfolded
	\item Sustained flight with two arms folded, allowing for the traversal of narrow tunnels
	\item Traversal of narrow gaps in projectile motion with all four arms folded
	\item Perching on wires with all four arms folded
	\item Grasping of, and flight with, light-weight objects of appropriate dimensions
\end{enumerate}

Note that, compared to \cite{bucki2019design}, the vehicle proposed in this paper is uniquely able to perform tasks (2) and (5).
Whereas in \cite{bucki2019design} the vehicle could only traverse narrow gaps in projectile motion with all four arms folded, in this work the vehicle is capable of hovering with two arms folded, which allows the vehicle to traverse narrow horizontal tunnels of arbitrary length and perform limited grasping tasks.
Additionally, in this paper we demonstrate the ability of the vehicle to perform task (4) for the first time, although the vehicle in \cite{bucki2019design} is theoretically capable of accomplishing this task as well.

By avoiding the use of complex mechanisms or additional actuators beyond the four motors used to drive the propellers, the proposed vehicle is capable of flying in the unfolded configuration with an efficiency nearly identical to that of a similarly designed conventional quadcopter.
The main drawback of our design (compared to a conventional quadcopter) is the fact that additional bounds must be placed on the four thrust forces such that the vehicle remains in the desired configuration during flight.
As we will show in Sections \ref{sec:vehicleAgility} and \ref{sec:figEight}, these bounds primarily reduce the maximum yaw torque that can be produced in the unfolded configuration, but do not otherwise significantly affect the agility of the vehicle.
Although the vehicle controller design is marginally more complex than that of a conventional quadcopter, it has a practically identical computational complexity, and can easily be implemented on existing flight controllers.

Furthermore, we argue that the mechanical simplicity of the proposed vehicle is its primary advantage when compared to other novel quadcopter-like designs capable of aerial morphing and/or grasping as described in Section \ref{sec:relWork}.
Because no additional actuators or complex mechanisms are required to perform morphing or grasping tasks, the proposed vehicle has both a lower cost (as no additional actuators must be purchased) and a longer flight time in the unfolded configuration (as no additional actuators must be carried) than other designs capable of morphing and/or grasping.
Conversely, the lack of additional actuators means that the proposed vehicle may not necessarily perform tasks that require grasping or morphing as well.
For example, the design presented in \cite{zhao2018deformable} (as well as most other designs) allow for the thrust vectors of all four propellers to remain parallel to gravity while grasping an object, allowing for heavier objects to be lifted than can be lifted using the design presented in this paper.
Additionally, such designs are capable of more efficient flight while grasping objects (or otherwise morphed) compared to the proposed vehicle in the two-arms-folded configuration, as all four arms can be used to lift the vehicle. 

Thus, we suggest that the proposed design is advantageous to existing morphing/grasping solutions in situations where the vehicle is expected to function primarily as a conventional quadcopter, i.e. when the vehicle is only occasionally required to morph/grasp.
For example, such use-cases are common in search and rescue applications where a vehicle must morph to traverse a narrow passageway (e.g. a window or doorway), but can return to normal flight operations upon traversing the passageway.
The proposed design is primarily useful in such situations because (aside from being lower cost) it is able to attain longer flight times in the unfolded configuration due to its lower mass, as no additional actuators or complex mechanisms are required.

%% file: sysModel.tex
\section{System model} \label{sec:sysModel}
In this section we follow \cite{bucki2019design} in defining a model of the system and deriving the dynamics of the vehicle.
The dynamics of the vehicle are then used in Section \ref{sec:contorl} to derive bounds on the control inputs such that the vehicle remains in the desired configuration.

\begin{figure}
	\includegraphics[width=\columnwidth]{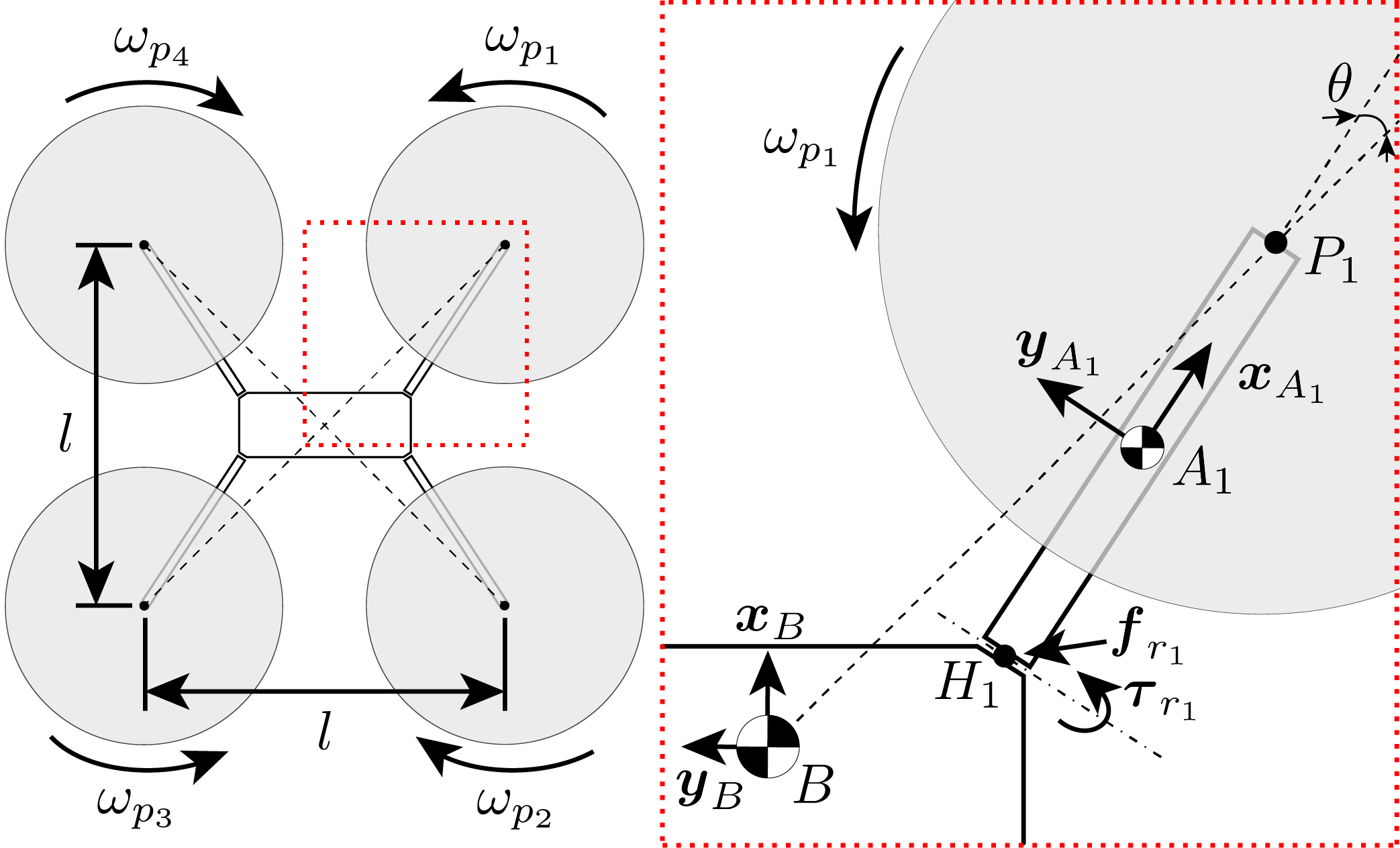}
	\centering
	\caption{Top-down view of vehicle in the unfolded configuration (left) and detail view of arm $A_1$ (right). In the unfolded configuration, the thrust axis of each rotor is parallel and equidistant from its neighbor, as is typical for quadcopters. Each arm is connected to the central body by a hinge that rotates in the \y{A_i}{} direction, allowing the arms to independently rotate between the folded and unfolded configurations. Internal forces $\force{r_i}{}$ and torques $\torque{r_i}{}$ act in equal and opposite directions between each arm and central body at $H_i$. The orientation of each hinge relative to the central body is determined by \armAngle. Each propeller produces a thrust force \thrust{i} and torque \propTorque{i} at $P_i$ in the direction of \z{A_i}{}.}
	\label{fig:sysModel}
\end{figure}

The vehicle consists of four rigid arms connected to a central body via unactuated rotary joints (i.e. hinges) which are limited to a range of motion of \SI{90}{\degree}.
Unlike \cite{bucki2019design}, each hinge is positioned such that the vehicle is only \SI{180}{\degree} axis-symmetric rather than \SI{90}{\degree} axis-symmetric (i.e. somewhat more like an ``H" than an ``X").
Figure \ref{fig:sysModel} shows a top-down view of the vehicle, including the orientation of each of the hinges and arms.

\subsection{Notation}
Non-bold symbols such as \mass{} represent scalars, lowercase bold symbols such as $\bs g$ represent first order tensors (vectors), and uppercase bold symbols such as \inertia{}{} represent second order tensors (matrices).
Subscripts such as \mass{B}{} represent the body to which the symbol refers, and superscripts such as $\bs g^E$ represent the frame in which the tensor is expressed.
A second subscript or superscript such as \angvel{BE}{} or \rot{BE} represents what the quantity is defined with respect to.
The symbol \dist{}{} represents a displacement, \angvel{}{} represents an angular velocity, and \rot{}{} represents a rotation matrix.
The skew-symmetric matrix form of the cross product is written as $\skewSymm{\bs a}$ such that $\skewSymm{\bs a}\bs b = \bs a \times \bs b$.

\subsection{Model}
The system is modeled as five coupled rigid bodies: the four arms and the central body of the vehicle.
The inertial frame is notated as $E$, the frame fixed to the central body as $B$, and the frame fixed to arm $i \in \curlyB{1,2,3,4}$ as $A_i$.
The rotation matrix of frame $B$ with respect to frame $E$ is defined as \rot{BE} such that the quantity $\bs v^B$ expressed in the $B$ frame is equal to $\rot{BE}\bs v^E$ where $\bs v^E$ is the same quantity expressed in frame $E$.
The orientation of arm $i$ with respect to the central body is defined through the single degree of freedom rotation matrix \rot{A_iB}.

When used in a subscript of a displacement tensor or its time derivatives, $E$ is defined as a fixed point in the inertial frame, $B$ as the center of mass of the central body, and $A_i$ as the center of mass of arm $i$.
For example, \dist{A_iB}{B} represents the displacement of the center of mass of arm $i$ with respect to the center of mass of the central body, and is expressed in the body-fixed frame $B$.
Furthermore, let $P_i$ be a point along the thrust axis of propeller $i$, and let $H_i$ be the point where the rotation axis of hinge $i$ intersects with the plane swept by the thrust axis of propeller $i$ as arm $i$ rotates about its hinge.

The internal reaction forces and torques acting at the hinge are defined as \force{r_i}{} and \torque{r_i}{} respectively.
The propeller attached to arm $i$ produces scalar thrust force \thrust{i} and aerodynamic reaction torque \propTorque{i} in the \z{A_i}{} direction.
We assume that the torque produced by each propeller is piecewise linearly related to the propeller thrust force \cite{pounds2002design} with positive proportionality constants \thrustToTorqueForward and \thrustToTorqueBack such that:
\begin{equation}\label{eq:thrustToTorque}
\propTorque{i} = 
\left\{ \begin{array}{ll}
(-1)^i\thrustToTorqueForward\thrust{i} & \thrust{i} \geq 0 \\
(-1)^i\thrustToTorqueBack\thrust{i} & \thrust{i} < 0
\end{array} \right.
\end{equation}
where $(-1)^i$ models the handedness of the propellers, $\thrust{i} < 0$ when the propellers are spun in reverse, and $\thrustToTorqueForward \neq \thrustToTorqueBack$ when asymmetric propellers are used, as is common with quadcopters.

The mass and mass moment of inertia of the central body taken at the center of mass of the central body are denoted \mass{B} and \inertia{B}{} respectively, and the mass and mass moment of inertia of arm $i$ taken at its center of mass are denoted \mass{A_i} and \inertia{A_i}{} respectively.

\subsection{Dynamics}
The translational and rotational dynamics of the central body of the vehicle and the four arms are found using Newton's second law and Euler's law respectively \cite{zipfel}.
We assume that the only external forces and torques acting on the vehicle are those due to gravity and the thrusts and torques produced by each propeller (for example, aerodynamic effects acting on the central body or arms are not considered).
The time derivative of a vector is taken in the reference frame in which that vector is expressed.
For more information regarding the derivation of the dynamics of coupled rigid bodies, see \cite{zipfel}.

We express the translational dynamics of the central body in the inertial frame $E$, and the rotational dynamics of the central body in the body-fixed frame $B$.
Let $\bs g$ be the acceleration due to gravity.
The translational dynamics of the central body are then:
\begin{equation}\label{eq:bodyTrans}
m_B\distddot{BE}{E} = m_B\bs g^E + \rot{EB}\sum_{i=1}^4 \force{r_i}{B}
\end{equation}
and the rotational dynamics of the central body are:
\begin{equation}
\begin{split}
\inertia{B}{B} &\angacc{BE}{B} + \skewSymm{\angvel{BE}{B}} \inertia{B}{B} \angvel{BE}{B} \\ 
&= \sum_{i=1}^4 \left(\torque{r_i}{B} + \skewSymm{\dist{H_iB}{B}}\force{r_i}{B}\right)
\end{split}
\end{equation} 

We express the translational and rotational dynamics of arm $i$ in frame $A_i$.
The translational dynamics of arm $i$ are (note $\force{r_i}{A_i} = \rot{A_iB}\force{r_i}{B}$):
\begin{equation}\label{eq:armTranslation}
m_{A_i}\left(\rot{A_iE}\distddot{BE}{E} + \bs \alpha\right) 
= m_{A_i}\rot{A_iE}\bs g^{E} + \z{A_i}{A_i}\thrust{i} - \force{r_i}{A_i}
\end{equation}
where $\bs \alpha$ is
\begin{equation}
\begin{split}
\bs \alpha = &\rot{A_iB}\left(\skewSymm{\dist{BH_i}{B}}\angacc{BE}{B} + \skewSymm{\angvel{BE}{B}}\skewSymm{\dist{BH_i}{B}}\angvel{BE}{B}\right) \\
&+ \skewSymm{\dist{H_iA_i}{A_i}}\angacc{A_iE}{A_i} + \skewSymm{\angvel{A_iE}{A_i}}\skewSymm{\dist{H_iA_i}{A_i}}\angvel{A_iE}{A_i} 
\end{split}
\end{equation}

The rotational dynamics of arm $i$ are (note $\torque{r_i}{A_i} = \rot{A_iB}\torque{r_i}{B}$):
\begin{equation}\label{eq:armRotation}
\begin{split}
\inertia{A_i}{A_i}&\angacc{A_iE}{A_i} + \skewSymm{\angvel{A_iE}{A_i}}\inertia{A_i}{A_i}\angvel{A_iE}{A_i} = \skewSymm{\dist{P_iA_i}{A_i}}\z{A_i}{A_i}\thrust{i} \\
&+ \z{A_i}{A_i}\propTorque{i} - \torque{r_i}{A_i} - \skewSymm{\dist{H_iA_i}{A_i}}\force{r_i}{A_i}
\end{split}
\end{equation}

The equations of motion of the arm are written in terms of $\angacc{A_iE}{A_i}$ and $\angvel{A_iE}{A_i}$ for convenience, which evaluate to:
\begin{equation}
\begin{split}
\angvel{A_iE}{A_i} &= \angvel{A_iB}{A_i} + \rot{A_iB}\angvel{BE}{B} \\
\angacc{A_iE}{A_i} &= \angacc{A_iB}{A_i} + \rot{A_iB} \angacc{BE}{B} - \skewSymm{\angvel{A_iB}{A_i}}\rot{A_iB}\angvel{BE}{B}
\end{split}
\end{equation}

Furthermore, note that the reaction torque acting in the rotation direction of hinge $i$ is zero when arm $i$ is rotating between the folded and unfolded configurations ($\y{A_i}{A_i}\cdot \torque{r_i}{A_i} = 0$), positive when arm $i$ is in the folded configuration ($\y{A_i}{A_i}\cdot \torque{r_i}{A_i} \geq 0$), and negative when arm $i$ is in the unfolded configuration ($\y{A_i}{A_i}\cdot \torque{r_i}{A_i} \leq 0$).
Thus, in order for arm $i$ to remain in a desired position when starting in that position (i.e. folded or unfolded), the vehicle must be controlled such that $\y{A_i}{A_i}\cdot \torque{r_i}{A_i}$ remains either positive (to remain folded) or negative (to remain unfolded).
Such a method is presented in the following section.

%% file: control.tex
\section{Control} \label{sec:contorl}

In this section we describe the controllers used to control the vehicle in each of its three configurations: the unfolded configuration (shown in Figure \ref{subfig:unfolded}), the two-arms-folded configuration (shown in Figures \ref{subfig:tunnel} and \ref{subfig:box}), and the four-arms-folded configuration (shown in Figure \ref{subfig:projectile}).

The vehicle is capable of controlled hover in both the unfolded and two-arms-folded configurations.
In the unfolded configuration, the vehicle acts as a conventional quadcopter; each of the four propellers produce positive thrust forces ($\thrust{i} > 0$) in the \z{B}{} direction.
However, in the two-arms-folded configuration, only two propellers of the same handedness produce positive thrust forces in the \z{B}{} direction; the other two propellers spin in reverse, producing negative thrust forces that cause their associated arms to fold downward.
In this configuration, the folded arms are positioned such that the thrust forces produced by their associated propellers create a yaw torque that counteracts the yaw torque produced by the other two propellers.
Note that for the design considered in this paper, the arms have a \SI{90}{\degree} range of motion such that the thrust produced by a folded arm has no component in the \z{B}{} direction.

In the four-arms-folded configuration each of the four propellers are spun in reverse ($\thrust{i} < 0$), resulting in all four arms folding.
Although the vehicle is not capable of sustained hover in this configuration, the attitude of the vehicle can still be fully controlled, allowing for the vehicle to reorient itself while in projectile motion.
This capability allows the vehicle to maintain a desired attitude while traversing narrow gaps in the four-arms-folded configuration, minimizing attitude errors when returning to the unfolded configuration, allowing for a more rapid recovery.

\begin{figure}
	\includegraphics[width=\columnwidth]{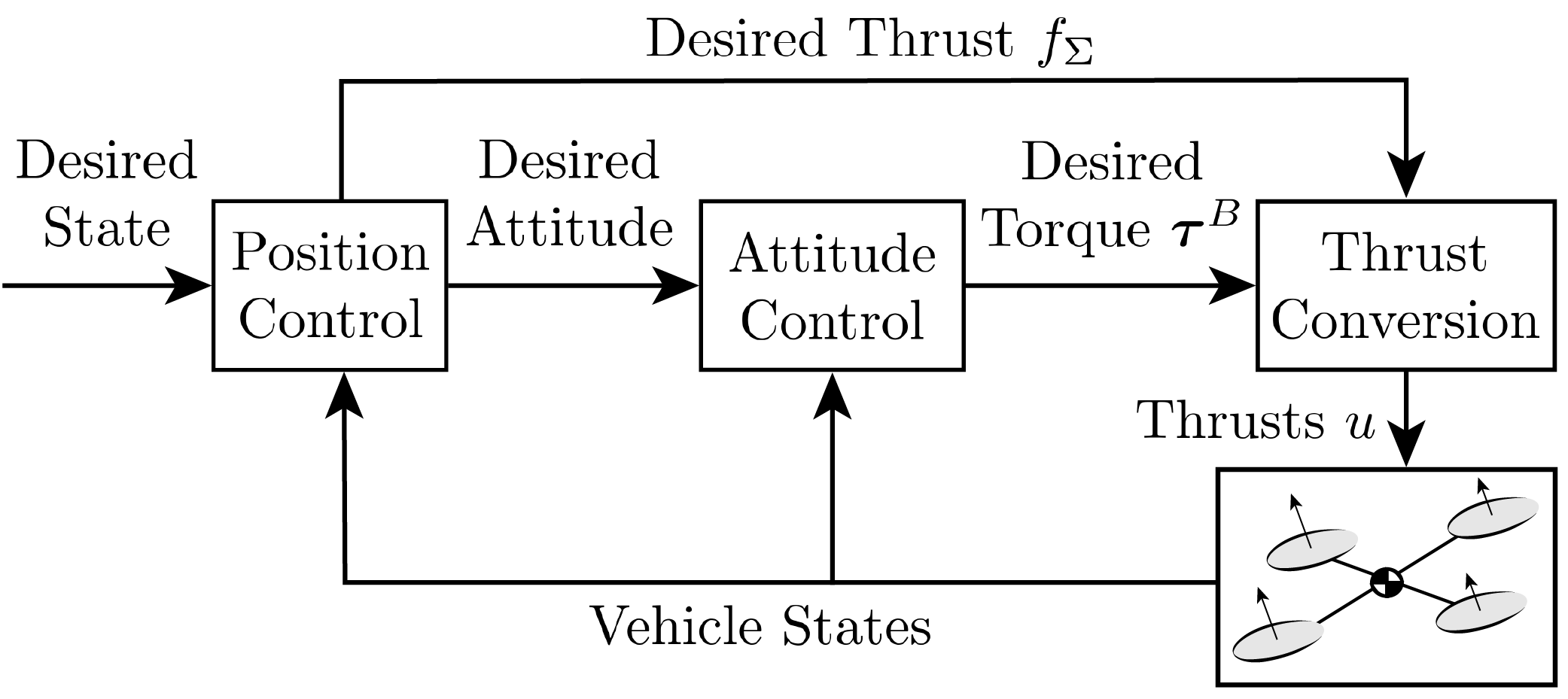}
	\centering
	\caption{Cascaded controller used to control the vehicle.}
	\label{fig:control}
\end{figure}

A cascaded control structure typical of multicopter control, shown in Figure \ref{fig:control}, is used to control the vehicle in both the unfolded and two-arms-folded configurations.
A position controller first computes a desired acceleration based on position and velocity errors, allowing for the computation of the desired thrust direction \z{B}{} and total thrust $\totalThrust$ in that direction.
Then, an attitude controller computes the desired torque required to align the thrust direction \z{B}{} with the desired acceleration direction and achieve the desired yaw angle.
Finally, the individual propeller thrust forces necessary to generate the desired total thrust and desired body torque are computed.
For each propeller, the desired thrust is converted to a desired angular velocity, which an electronic speed controller tracks.

A similar control structure is used in the four-arms-folded configuration, but with a slight modification.
Because thrust cannot be produced in the \z{B}{} direction, the position of the vehicle is not controlled in this configuration, but the attitude can be controlled.
Thus, in this configuration we omit the position controller and instead command desired attitudes to the attitude controller directly, allowing for the vehicle to reorient itself while undergoing projectile motion.
As the desired thrust $\totalThrust$ from the position controller is neglected in this configuration, the Moore-Penrose pseudoinverse of the matrix relating the four thrust forces to the torque acting on the vehicle is used to compute the minimum 2-norm of the thrust forces necessary to achieve the desired torques.
Note that more sophisticated methods (e.g. solving a convex optimization problem) could also be used to convert the desired torques into the four thrust forces.

Although there exist many different control structures for controlling multicopters, we choose the cascaded control structure shown in Figure \ref{fig:control} primarily due to its simplicity and widespread use in controlling conventional quadcopters \cite{mahony2012multirotor}.
The primary difference between the controller presented in this work and existing cascaded controllers is the use of a bound on the desired total thrust $\totalThrust$ and desired torque $\torque{}{B}$ that prevents the arms from folding or unfolding when the vehicle is not changing configurations.
Although we provide details regarding the design of an attitude controller for the vehicle in Section \ref{sec:attCtrl}, we do not consider the attitude controller a novel contribution, as similar attitude control methods have been applied in previous works (e.g. \cite{bucki2019novel}).
Furthermore, note that this choice of control structure does not preclude other nonlinear controllers from being used to control the proposed vehicle; rather, this choice of a cascaded controller is made in order to provide a simple example of a method of controlling the proposed vehicle.

\subsection{Individual thrust force computation}\label{sec:indivThrustComp}
The individual propeller thrust forces $\thrusts = (\thrust{1}, \thrust{2}, \thrust{3}, \thrust{4})$ are related to the desired total thrust in the \z{B}{} direction $\totalThrust$ and the desired torques about the axes of the body-fixed $B$ frame, $\bodyTorque = (\tau_x, \tau_y, \tau_z)$ as follows: 
\begin{equation}\label{eq:mappingDef}
\begin{bmatrix}
\totalThrust \\ \bodyTorque
\end{bmatrix} =
\begin{bmatrix}
M_{\totalThrust} \\
M_{\bodyTorque}
\end{bmatrix}
\thrusts = M\thrusts
\end{equation}
where $M_{\totalThrust} \in \mathbb{R}^{1\times 4}$ is the mapping from $\thrusts$ to $\totalThrust$, $M_{\bodyTorque} \in \mathbb{R}^{3\times 4}$ is the mapping from $\thrusts$ to $\bodyTorque$, and $M \in \mathbb{R}^{4\times 4}$ is the combined mapping.

The mapping $M$ is computed using the geometry of the vehicle and the torque produced by each propeller as a function of the thrust it produces.
Let \dist{P_iC}{} be the position of propeller $P_i$ relative to the center of mass of the entire vehicle $C$, and $\thrustToTorque{i} = (-1)^i\thrustToTorqueForward$ or $\thrustToTorque{i} = (-1)^i \thrustToTorqueBack$ depending on the thrust direction of propeller $i$ as defined in \eqref{eq:thrustToTorque}.
Then, the $i$-th columns of $M_{\totalThrust}$ and $M_{\bodyTorque}$ are
\begin{equation}
M_{\totalThrust}[i] = \z{A_i}{B}\cdot \z{B}{B}, \quad M_{\bodyTorque}[i] = \skewSymm{\dist{P_iC}{B}}\z{A_i}{B} + \thrustToTorque{i}\z{A_i}{B}
\end{equation}
where we recall that $\z{A_i}{B}$ is a unit vector written in the body-fixed frame $B$ that points in the positive thrust direction of propeller $i$.

Thus, in the unfolded configuration, $\unfoldedMixer$ is the mapping of a typical quadcopter with $l$ defined as shown in Figure \ref{fig:sysModel}:
\begin{equation}\label{eq:unfoldedMmixer}
\unfoldedMixer = \begin{bmatrix}
1 & 1 & 1 & 1 \\
-l/2 & -l/2 & l/2 & l/2 \\
-l/2 & l/2 & l/2 & -l/2 \\
-\thrustToTorqueForward & \thrustToTorqueForward & -\thrustToTorqueForward & \thrustToTorqueForward
\end{bmatrix}
\end{equation}

The mapping $\twoArmsMixer$ for the two-arms-folded configuration with arms $A_2$ and $A_4$ folded and $\armAngle$ defined as shown in Figure \ref{fig:sysModel} is defined as follows.
An equivalent mapping exists for the two-arms-folded configuration with arms $A_1$ and $A_3$ folded.
\begin{equation}\label{eq:twoArmsFoldedMixer}
\begin{gathered}
\twoArmsMixer = \begin{bmatrix}
1 & 0 & 1 & 0 \\
-l/2 & p_x & l/2 & -p_x \\
-l/2 & -p_y & l/2 & p_y \\
-\thrustToTorqueForward & -p_z & -\thrustToTorqueForward & -p_z
\end{bmatrix} \\
p_{x} = \distz{P_2C}{B}\cos(\SI{45}{\degree} + \armAngle) - \thrustToTorqueBack\sin(\SI{45}{\degree} + \armAngle) \\
p_{y} = \distz{P_2C}{B}\sin(\SI{45}{\degree} + \armAngle) + \thrustToTorqueBack\cos(\SI{45}{\degree} + \armAngle) \\
p_z = \frac{l\sqrt{2}}{2}\sin\armAngle
\end{gathered}
\end{equation}
where we note that the arms are of equal length, i.e. $\distz{P_1C}{B} = \distz{P_2C}{B} = \distz{P_3C}{B} = \distz{P_4C}{B}$.

Finally, the mapping $\fourArmsMixer$ for the four-arms folded configuration is defined as:
\begin{equation}\label{eq:fourArmsFoldedMixer}
\fourArmsMixer = \begin{bmatrix}
0 & 0 & 0 & 0 \\
p_x & p_x & -p_x & -p_x \\
p_y & -p_y & -p_y & p_y \\
p_z & -p_z & p_z & -p_z
\end{bmatrix}
\end{equation}

The structure of these mappings can be analyzed to infer how different parameters of the vehicle affect how the vehicle can be controlled in each configuration.
For example, we note that when the arms are not angled (i.e. when $\armAngle = 0$ as was done in our prior work \cite{bucki2019design}), the term $p_z$ as defined in \eqref{eq:twoArmsFoldedMixer} equals zero.
In this case, the vehicle would only be able to produce negative yaw torques due to the fact that the folded arms would be unable to offset the yaw torque produced by the two unfolded arms.
Similarly, if $\armAngle = 0$, the vehicle would be unable to produce any yaw torque in the four-arms-folded configuration as the bottom row of $\fourArmsMixer$ would be zero.
However, we observe that \eqref{eq:unfoldedMmixer} does not depend on $\armAngle$ at all, showing that the thrust mapping matrix when flying in the unfolded configuration is unaffected by the choice of arm angle.

Note that, unlike $\unfoldedMixer$, both $\twoArmsMixer$ and $\fourArmsMixer$ depend on the position of the center of mass of the vehicle in the \z{B}{} direction due to the fact that the thrust forces produced by the folded arms are perpendicular to \z{B}{}.
Thus, if the position of the center of mass of the vehicle in the \z{B}{} direction is changed (e.g. by adding a payload to the vehicle as shown in Figure \ref{subfig:box}), the mappings $\twoArmsMixer$ and $\fourArmsMixer$ must reflect this change.

Furthermore, because the thrust forces of the folded arms are perpendicular to \z{B}{}, there can exist a nonzero force in the \x{B}{} and \y{B}{} directions when flying in the two- or four-arms folded configurations.
In the two-arms-folded configuration with arms $A_2$ and $A_4$ folded, for example, thrusts $\thrust{2}$ and $\thrust{4}$ act in opposite directions such that they produce a force of magnitude $|\thrust{2} - \thrust{4}|$.
Because this force is zero at hover and remains small for small $\tau_x$ and $\tau_y$ (note that $|\thrust{2} - \thrust{4}|$ is not dependent on $\totalThrust$ or $\tau_z$ due to the structure of $\twoArmsMixer$), we choose to treat such forces as unmodeled disturbances in order to maintain the simplicity of the proposed controller.

Note that a disturbance observer could be included in the proposed controller in order to estimate e.g. the difference in thrusts $\thrust{2}$ and $\thrust{4}$ when in the two-arms-folded configuration or e.g. wind disturbances when in any configuration.
Alternately, a more complex controller could be used that explicitly accounts for the lateral acceleration of the vehicle due to the difference in thrusts $\thrust{2}$ and $\thrust{4}$, for example.
However, we omit such extensions both for the sake of brevity and because acceptable performance was achieved on the experimental vehicle without using such methods.
Nonetheless, because we use a cascaded controller similar to those used with conventional quadcopters, existing disturbance observers such as \cite{dong2014high} can be used with the proposed vehicle without significant modification.


\subsection{Attitude control}\label{sec:attCtrl}
The attitude controller is designed using desired first-order behavior, described here by the rotation vector $\attError = (\phi_e, \theta_e, \psi_e)$ that represents the rotation between the current and desired attitude (i.e. a rotation about the axis defined in the direction of $\attError$ by angle $||\attError||$).
Note that, to first order, $\phi_e$, $\theta_e$, and $\psi_e$ represent roll, pitch, and yaw respectively.
The desired attitude is defined as that attitude at which the yaw angle of the vehicle matches the desired yaw angle and at which the thrust direction of the vehicle matches the desired thrust direction, which is given by the position controller (see Figure \ref{fig:control}).

The linearized attitude dynamics of the vehicle are then
\begin{equation}
\begin{bmatrix}
\dot{\attError} \\
\ddot{\attError}
\end{bmatrix} = 
A
\begin{bmatrix}
\attError \\
\dot{\attError}
\end{bmatrix} + 
B\bodyTorque
\end{equation}
where
\begin{equation}
A = \begin{bmatrix}
\bs 0 & \bs I \\
\bs 0 & \bs 0
\end{bmatrix}, \quad
B = \begin{bmatrix}
\bs 0 \\
\roundB{\inertia{C}{C}}^{-1}
\end{bmatrix}
\end{equation}
and where $\inertia{C}{C}$ is the moment of inertia of the entire vehicle written at its center of mass.
Note that $\inertia{C}{C}$ depends on the configuration of the vehicle and can be computed analytically from $\inertia{A_i}{A_i}$ and $\inertia{B}{B}$ using the parallel-axis theorem, or (e.g. for the experimental vehicle presented in Section \ref{sec:design}) using computer-aided design software.

We choose to synthesize different infinite-horizon LQR controllers \cite{LQR} to control the attitude of the vehicle when flying in different configurations.
Although other control strategies (e.g. model predictive control) could be used, we choose to use an LQR controller because it is both reasonably simple and can be run on low-power embedded devices with minimal setup (i.e. it does not require any specialized hardware to be used or for any optimization problems to be solved online).
The LQR controller is synthesized using state cost matrix $\stateCost \in \mathbb{R}^{6\times 6}$ and input cost matrix $\inputCost \in \mathbb{R}^{3\times 3}$.
For each configuration of the vehicle, we weigh the cost of each state error independently such that $\stateCost$ is a diagonal matrix.
The values of the diagonal of $\stateCost$ are chosen such that the costs associated with $\phi_e$ and $\theta_e$ (i.e. elements 1 and 2) are equal and such that the costs associated with the roll rate and pitch rate (i.e. elements 4 and 5) are equal.
We define the input cost matrix $\inputCost$ using the mapping $M_{\bodyTorque}$ from the individual thrust forces $\thrusts$ to the desired torque $\bodyTorque$ as defined in \eqref{eq:mappingDef} (i.e. the lower three rows of $\unfoldedMixer$, $\twoArmsMixer$, or $\fourArmsMixer$, depending on the configuration of the vehicle):
\begin{equation}\label{eq:inputCost}
\inputCost = (M_{\bodyTorque}^{+})^T \inputCostThrusts M_{\bodyTorque}^{+}
\end{equation}
where $M_{\bodyTorque}^{+}$ is the pseudoinverse of the mapping matrix $M_{\bodyTorque}$, and $\inputCostThrusts \in \mathbb{R}^{4\times 4}$ is a diagonal matrix that encodes the cost associated with the thrust force produced by each propeller.

In this work, we choose the diagonal entries of $\inputCostThrusts$ based upon whether the associated propeller is spinning in the forward or reverse direction, as the propeller exhibits different characteristics in each mode of operation.
For example, conventional propellers produce significantly less thrust when spinning in the reverse direction as they are typically optimized to spin in only the forward direction.
This differs from our previous works which use similar attitude controllers \cite{bucki2019design} and \cite{bucki2019novel}, where the propellers were constrained to only spin in the forward direction.
Thus, we define $\inputCostThrusts = \textrm{diag}(\forwardThrustCost, \forwardThrustCost, \forwardThrustCost, \forwardThrustCost)$ for the unfolded configuration, $\inputCostThrusts = \textrm{diag}(\forwardThrustCost, \reverseThrustCost, \forwardThrustCost, \reverseThrustCost)$ for the two-arms-folded configuration, and $\inputCostThrusts = \textrm{diag}(\reverseThrustCost, \reverseThrustCost, \reverseThrustCost, \reverseThrustCost)$ for the four-arms-folded configuration, where $\forwardThrustCost$ is the cost associated with the propellers spinning in the forward direction, and $\reverseThrustCost$ is the cost associated with the propellers spinning in the reverse direction.
In general, $\forwardThrustCost < \reverseThrustCost$ as conventional quadcopter propellers are optimized to spin in the forward direction.

By defining the input cost matrix $\inputCost$ as a function of the mapping matrix $M_{\bodyTorque}$, we can straightforwardly synthesize different infinite-horizon LQR attitude controllers for each configuration of the vehicle.
Furthermore, the torque cost matrix $\inputCost$ can be used to analyze the ability of the vehicle to control its attitude in different configurations, as it describes the cost of producing an arbitrary torque on the vehicle while implicitly accounting for the geometry of the vehicle due to its dependence on $M_{\bodyTorque}$.



\subsection{Thrust limits}\label{sec:thrustLimits}

Although the thrust produced by each propeller is already bounded by the performance limitations of the motor driving it, we impose additional bounds which ensure the vehicle remains in the desired configuration.
Imposing these bounds ensures that none of the arms begin to fold or unfold unexpectedly, which means the mappings $\unfoldedMixer$, $\twoArmsMixer$, and $\fourArmsMixer$ derived in Section \ref{sec:indivThrustComp} will remain valid during flight.
Of course, the bounds are not imposed when changing between configurations.

Rather than bounding the individual thrust forces, we choose to instead bound $\totalThrust$ and $\bodyTorque$ using the model derived in Section \ref{sec:sysModel}.
Our approach is similar to that of \cite{bucki2019design}, but differs in its inclusion of the arm angle \armAngle, resulting in a modified expression for the bound.
We choose to write these bounds in terms of $\totalThrust$ and $\bodyTorque$ in order to allow for a hierarchical modification of $\totalThrust$ and $\bodyTorque$ when the bound is not satisfied which prioritizes the roll and pitch torques over the yaw torque and total thrust produced by the propellers (similar to \cite{faessler2017thrust}).
This prioritization is chosen based on the cascaded control structure described previously, where the roll and pitch torques computed by the attitude controller are used to quickly align the vehicle's thrust direction with the desired acceleration direction computed by the position controller.
We prioritize the acceleration direction over yaw, as it corresponds to the safety critical position control.

\subsubsection{Unfolded configuration bounds}
We first note that by enforcing bounds that prevent each arm from folding or unfolding, the vehicle can be treated as one rigid body rather than five coupled rigid bodies.
Thus, the acceleration of the center of mass of the vehicle expressed in the inertial frame $\distddot{CE}{E}$ is:
\begin{equation}\label{eq:accCE}
\distddot{CE}{E} = \bs g^E + \frac{1}{m_\Sigma}\rot{EB}\z{B}{B}\totalThrust
\end{equation}
where the total vehicle mass is $m_\Sigma = m_B + 4 m_{A_i}$.

Similarly, the angular acceleration of the vehicle can be written as follows, where $\totalInertia$ represents the moment of inertia of the vehicle taken at its center of mass and expressed in the body-fixed frame $B$.
We assume that the angular velocity of the vehicle $\angvel{BE}{B}$ is small such that second order terms with respect to $\angvel{BE}{B}$ can be neglected (e.g. $\skewSymm{\angvel{BE}{B}}\totalInertia\angvel{BE}{B}$).
Thus, the derived bound may have some error when performing e.g. acrobatic maneuvers that result large magnitude second order terms.
These effects may be compensated for by including an additional factor of safety in the derived bound, but in practice we have found this to be unnecessary, as the second order terms are negligible compared to the other terms in the bound (e.g. during flight near hover and simple trajectories).
Therefore, by including this assumption,
\begin{equation}\label{eq:angaccCE}
\angacc{BE}{B} \approx \roundB{\totalInertia}^{-1}\bodyTorque
\end{equation}

Next, after some algebraic manipulation of \eqref{eq:armTranslation} and \eqref{eq:armRotation} (omitted here for brevity), we find that the reaction torque about hinge $i$, i.e. $\y{A_i}{A_i}\cdot \torque{r_i}{A_i}$, is linear with respect to $\distddot{CE}{E}$, $\angacc{BE}{B}$, and propeller thrust $\thrust{i}$.
Recall that $\thrust{i}$ can be computed by inverting the mapping given in \eqref{eq:mappingDef}, meaning that $\distddot{CE}{E}$, $\angacc{BE}{B}$, and $\thrust{i}$ are all linear functions of $\totalThrust$ and $\bodyTorque$.
Thus, we find that the torque about hinge $i$ is also a linear function of $\totalThrust$ and $\bodyTorque$.

Arm $i$ will remain in the unfolded configuration when $\y{A_i}{A_i}\cdot \torque{r_i}{A_i} \leq 0$.
Therefore, because $\y{A_i}{A_i}\cdot \torque{r_i}{A_i}$ is linear with respect to $\totalThrust$ and $\bodyTorque$, the following four bounds can be computed that ensure each of the four arms remain in the unfolded configuration:
\begin{equation}\label{eq:thrustLimitPerArm}
c_{f_i}\totalThrust + c_{x_i}\tau_x + c_{y_i}\tau_y + c_{z_i}\tau_z \geq 0,\quad i\in \{1,2,3,4\}
\end{equation}
where $c_{f_i}$, $c_{x_i}$, $c_{y_i}$, and $c_{z_i}$ are constants that depend on the physical attributes of the vehicle.

For the unfolded configuration, the constants in \eqref{eq:thrustLimitPerArm} are as follows.
Here we have included the assumption that $\totalInertia = \textrm{diag}(\totalInertiaxx, \totalInertiayy, \totalInertiazz)$ in order to allow for clearer analysis of $c_{x_i}$, $c_{y_i}$, and $c_{z_i}$.
We give the magnitudes of each term, noting that $c_{x_i}$, $c_{y_i}$, and $c_{z_i}$ have different signs depending which arm they are associated with.
\begin{equation}\label{eq:cf}
c_{f_i} = \frac{1}{4}\distx{P_iH_i}{A_i} - \distx{A_iH_i}{A_i}\frac{\mass{A_i}}{m_\Sigma}
\end{equation}
\begin{equation}
\absB{c_{x_i}} = \absB{\frac{\distx{P_iH_i}{A_i}}{2l} - \frac{\widetilde{J}_{A_i,yy}^{A_i}\cos(\SI{45}{\degree}\!\! +\!\armAngle) + \tilde{m}_A\sin(\SI{45}{\degree}\!\! +\!\armAngle)}{\totalInertiaxx}}
\end{equation}
\begin{equation}
\absB{c_{y_i}} = \absB{-\frac{\distx{P_iH_i}{A_i}}{2l} + \frac{\widetilde{J}_{A_i,yy}^{A_i}\sin(\SI{45}{\degree}\!\! + \!\armAngle) + \tilde{m}_A\cos(\SI{45}{\degree}\!\! + \!\armAngle)}{\totalInertiayy}}
\end{equation}
\begin{equation}\label{eq:cz}
\absB{c_{z_i}} = \absB{\frac{\distx{P_iH_i}{A_i}}{4\thrustToTorqueForward} - \frac{\tilde{m}_A}{\totalInertiazz}}
\end{equation}
where $\tilde{m}_A$ and $\tilde{J}_{A_i,yy}^{A_i}$ are
\begin{equation}
\tilde{J}_{A_i,yy}^{A_i} = \roundB{\inertia{A_i}{A_i} + \mass{A_i}\skewSymm{\dist{A_iH_i}{A_i}}\skewSymm{\dist{CA_1}{A_1}}}_{yy}
\end{equation}
\begin{equation}
\tilde{m}_A = \mass{A_i}\distx{A_iH_i}{A_i}\disty{CH_1}{A_1}
\end{equation}

Because of the equal magnitudes of the constants $c_{f_i}$, $c_{x_i}$, $c_{y_i}$, and $c_{z_i}$ in the unfolded configuration, we can aggregate the four bounds given in \eqref{eq:thrustLimitPerArm} into a single bound:
\begin{equation}\label{eq:aggregateThrustBound}
c_{f_i}\totalThrust - |c_{x_i}\tau_x| - |c_{y_i}\tau_y| - |c_{z_i}\tau_z| \geq 0
\end{equation}

Note that \eqref{eq:aggregateThrustBound} can be satisfied by increasing $\totalThrust$, as this corresponds to requiring each propeller to produce more thrust (note that in general $c_{f_i} > 0$).
By examining \eqref{eq:cf}, we observe that the bound becomes less restrictive when, e.g., the ratio of the mass of an arm to the total mass of the vehicle decreases, as this results in a larger magnitude $c_{f_i}$.
Similarly, because the magnitude of $c_{z_i}$ decreases as $\thrustToTorqueForward$ increases, the bound can be made less restrictive by, e.g., choosing propellers with a larger magnitude $\thrustToTorqueForward$.

Finally, note that by writing this bound as a function of $\totalThrust$ and $\bodyTorque$, we can apply a similar method to that presented in \cite{faessler2017thrust} to reduce these control inputs in the event that the bound is not satisfied.
Specifically, if the controller presented in the previous subsections produces a $\totalThrust$ and $\bodyTorque$ that does not satisfy \eqref{eq:aggregateThrustBound}, we first reduce the magnitude of the yaw torque $\tau_z$ until the bound is satisfied or $\tau_z = 0$.
Next, if the bound is still not satisfied, we increase $\totalThrust$ until the bound is satisfied or it reaches the maximum total thrust the propellers can produce.
If the maximum total thrust is reached, then the roll and/or pitch torques are reduced until the bound is satisfied.
In practice, however, decreasing the roll and/or pitch torques in order to prevent the arms from folding is seldom necessary due to the magnitude of $c_{x_i}$ and $c_{y_i}$ relative to the other terms.

\subsubsection{Two- and four-arms-folded configuration bounds}

Similar expressions for $c_{f_i}$, $c_{x_i}$, $c_{y_i}$, and $c_{z_i}$ can be found for the two- and four-arms-folded configurations, which we compute using a computer algebra system due to their algebraic complexity (and thus omit here for brevity).\footnote{We provide code for computing these bounds, as well as performing much of the other analyses and controller syntheses described in this paper, at \cite{github_repo}}
Note that no aggregate bound such as \eqref{eq:aggregateThrustBound} exists for the two- or four-arms-folded configuration, and thus it is necessary to enforce each bound given by \eqref{eq:thrustLimitPerArm} individually.
However, the hierarchical modification of the control inputs $\totalThrust$ and $\bodyTorque$ described previously can still be used to ensure the bounds are satisfied, guaranteeing that the vehicle remains in the desired configuration under the previously stated assumptions.

Numerical values for $c_{f_i}$, $c_{x_i}$, $c_{y_i}$, and $c_{z_i}$ are given in Section \ref{sec:vehicleAgility} for the experimental vehicle in both the unfolded and two-arms-folded configurations.
We do not provide such values for the four-arms-folded configuration, as in practice we have found it to be unnecessary to enforce such bounds.
This is because the thrust forces required to transition into the four-arms-folded configuration are typically large enough to prevent the arms from unfolding without the need to enforce additional bounds.


%
%

\subsection{Configuration transitions}\label{sec:transitions}

Next we describe the method used to transition between configurations of the vehicle.
We choose to focus on the transitions between the unfolded and two-arms-folded configurations as well as between the unfolded and four-arms-folded configurations, as these are the only transitions required to produce the behaviors of the vehicle demonstrated in this paper.
An example of the transition from the unfolded configuration to the two-arms-folded configuration and back is given in Section \ref{sec:tunnel}, and an example of the transition from the unfolded configuration to the four-arms-folded configuration and back is given in Section \ref{sec:vertGap}.

When transitioning between the unfolded and two-arms-folded configurations, we have found it sufficient to instantaneously change between the controller used in the unfolded configuration and the controller used in the two-arms-folded configuration.
This discrete change in controllers is largely enabled by the fact that the vehicle possesses sufficient agility in either configuration to recover from small disturbances encountered during the transition.
However, the transition is complicated by the fact that the vehicle experiences a significant yaw disturbance during the transition.
This yaw disturbance occurs because it is necessary to reverse the rotation direction of two of the propellers of the same handedness during the transition.
Specifically, the reversing propellers cannot offset the yaw torque produced by the propellers attached to the unfolded arms, which remain spinning in the forward direction.
Additionally, the reversing propellers experience a change in angular momentum that results in a corresponding change in angular velocity of the vehicle.
Thus, after completing the maneuver, the vehicle will have a significantly different yaw angle and yaw rate than when the maneuver was initiated. 
In practice, we deal with this difference in yaw angle by choosing the post-transition desired yaw angle such that once the maneuver is completed the yaw error is small.
Results demonstrating how these effects affect the experimental vehicle during configuration transitions can be found in Section \ref{sec:tunnel}. 

Unlike the transition to or from the two-arms-folded configuration, the transitions between the unfolded and four-arms-folded configurations are accomplished by commanding constant forward or reverse thrusts while the four arms are moving to the unfolded or folded configurations respectively.
After all four arms have finished transitioning, we resume controlling the vehicle using either the unfolded or two-arms-folded controller as appropriate.
The period of constant thrusts is required to ensure that all four arms fold or unfold simultaneously, and prevents any attitude errors that would otherwise be introduced by attempting to control the vehicle while the arms are transitioning (as $\unfoldedMixer$ and $\fourArmsMixer$ would not be valid during the transition).
Note that this method of transitioning to the four-arms-folded configuration differs from the method used in \cite{bucki2019design}, where the propellers were disabled and spring forces were used to fold the arms.
Because no spring forces are present in the proposed design, simply disabling all four motors would not result in the arms folding, as no forces would produce a relative acceleration between the four arms and the central body (aside from those produced by differences in the drag forces exerted on the arms and central body, which are typically too small to fold the arms in a reasonably short time, if at all).
Thus, reversing the thrust direction of all four propellers is required to transition into the four-arms-folded configuration.

%% file: design.tex
\section{Experimental Vehicle Design} \label{sec:design}
In this section, we discuss the design of the experimental vehicle shown in Figure \ref{fig:vehiclePic}.
We start by describing how the arm angle was chosen based upon other properties of the vehicle, then discuss how the properties of the chosen powertrain (i.e. the battery, speed controllers, motors, and propellers) affect the vehicle design, and finally discuss how the design of the vehicle influences several important properties of the proposed controllers for each configuration of the vehicle.

The properties of the experimental vehicle are given in Table~\ref{tab:vehicleParams}.
The overall dimensions of the experimental vehicle were chosen to be as similar as possible to a commonly used quadcopter design.
Specifically, \SI{8}{in} propellers spaced \SI{24}{\centi\meter} apart are used, which correspond to the same spacing and size of the propellers that would be used with a DJI F330 frame (e.g. as used in \cite{bucki2019novel}).
We chose to use commonly available components in the vehicle design to demonstrate its similarity to a conventional quadcopter, and designed the vehicle to have a similar performance (in terms of power consumption and agility) as a conventional quadcopter when flying in the unfolded configuration.

Because the hinges used to fold and unfold the arms may experience large dynamic loads during operation of the vehicle, they must be designed to withstand both loading experienced during flight, as well as the (often larger) impact forces applied when the arms of the vehicle move between the unfolded and folded configurations.
When hovering in the two-arms-folded configuration, the two unfolded arms must support the full weight of the vehicle, meaning that they experience twice the load as when flying in the unfolded configuration.
However, because the maximum thrust force of the two unfolded arms remains unchanged, the maximum load experienced by the arms in the two-arms-folded configuration is the same as in the unfolded configuration.
Note that the impact forces applied to the hinges during configuration transitions depend on several design parameters of the vehicle (e.g. the stiffness of materials which make contact when folding/unfolding), and as such may be difficult to estimate accurately.
For the experimental vehicle, the arms and central body were 3D printed using PLA, and were connected by free-rotating \SI{4}{\milli\meter} steel bolts.
We found this design to be capable of withstanding many repeated experiments, and observed no failures throughout our experiments.

Onboard the vehicle, a Crazyflie 2.0 flight controller is used to run the attitude controller and to transmit individual propeller angular velocity commands to four DYS SN30A electronic speed controllers (ESCs) at 500Hz.
The vehicle is powered by a three cell, 40C, \SI{1500}{\milli\ampere\hour} LiPo battery, and four EMAX MT2208 brushless motors are used to drive four Gemfan 8038 propellers.

A motion capture system is used to localize the vehicle, although in principle any sufficiently accurate localization method (e.g. using onboard cameras) could be used.
Note that we do not directly measure the position of any individual arm of the vehicle, and instead only measure the position and attitude of the central body of the vehicle.
The position and attitude of the vehicle are measured by the motion capture system at 200Hz, and the angular velocity of the vehicle is measured at 500Hz using an onboard rate gyroscope.
An extended Kalman filter is used to estimate the position, velocity, and attitude of the vehicle from the motion capture measurements, and a low-pass filter is used to estimated the angular velocity of the vehicle from the rate gyroscope measurements.
The position controller runs on an offboard laptop and sends commands to the vehicle via radio at 50Hz.
All control rates were chosen primarily based on hardware limitations, and were not found to limit the capabilities of the proposed vehicle.

The parameters used to compose the cost matrices $\stateCost$ and $\inputCostThrusts$ used to synthesize the LQR attitude controller described in Section \ref{sec:attCtrl} are given in Table~\ref{tab:vehicleControl}.
We define the attitude state cost matrix as $\stateCost = \textrm{diag}(\rollPitchCost, \rollPitchCost, \yawCost, \rollPitchRateCost, \rollPitchRateCost, \yawRateCost)$.
Higher costs are assigned to roll and pitch errors than yaw errors, as the cascaded control structure (show in Figure \ref{fig:control}) requires minimal errors between the vehicle thrust direction and the desired acceleration direction computed by the position controller.
The input cost matrix is computed according to \eqref{eq:inputCost}, where $\inputCostThrusts$ is determined by $\forwardThrustCost$, $\reverseThrustCost$, and the configuration of the vehicle as described in Section \ref{sec:attCtrl}.
We provide code for synthesizing the LQR attitude controller for each configuration in \cite{github_repo}.

\begin{table}[]
	\centering
	\caption{Experimental Vehicle Parameters}
	\label{tab:vehicleParams}
	\begin{tabular}{|c|c|c|}
		\hline
		\textbf{Symbol} & \textbf{Parameter} & \textbf{Value} \\ \hline
		$m_{A_i}$ &   Arm mass     &   \SI{67}{\gram}    \\ \hline
		$m_B$    &   Central body mass     &      \SI{356}{\gram} \\ \hline
		$m_\Sigma$ & Total vehicle mass & \SI{624}{\gram} \\ \hline
		$\thrustToTorqueForward$ & \makecell{Propeller torque per \\ unit positive thrust} & \SI{0.0172}{\newton\meter/\newton} \\ \hline
		$\thrustToTorqueBack$ & \makecell{Propeller torque per \\ unit negative thrust} & \SI{0.038}{\newton\meter/\newton} \\ \hline
		$\minThrust$ & Minimum thrust per propeller & \SI{-3.4}{\newton} \\ \hline
		$\maxThrust$ & Maximum thrust per propeller & \SI{7.8}{\newton} \\ \hline
		$\armAngle$ & Arm angle & \SI{11.9}{\degree} \\ \hline
		$l$ & \makecell{Distance between \\ adjacent propellers} & \SI{24}{\centi\meter} \\ \hline
		$\dist{BH_1}{B}$ & \makecell{Position of central body center \\ of mass relative to hinge 1 \\ (written in B frame)} & $\begin{pmatrix}
		\SI{-4.5}{\centi\meter}\\ \SI{7.1}{\centi\meter}\\ \SI{-0.2}{\centi\meter}
		\end{pmatrix}$ \\ \hline
		$\dist{H_iA_i}{A_i}$ & \makecell{Position of hinge relative \\ to arm center of mass \\ (written in arm frame)} & $\begin{pmatrix}
		\SI{-7.6}{\centi\meter}\\ \SI{0}{\centi\meter}\\ \SI{-1.4}{\centi\meter}
		\end{pmatrix}$ \\ \hline
		$\distx{PA_i}{A_i}$ & \makecell{Distance of propeller from \\ arm center of mass} & \SI{1.4}{\centi\meter} \\ \hline
	\end{tabular}
\end{table}

\begin{table}[]
	\centering
	\caption{Experimental Vehicle Controller Parameters}
	\label{tab:vehicleControl}
	\begin{tabular}{|c|c|c|}
		\hline
		\textbf{Symbol} & \textbf{Parameter} & \textbf{Value} \\ \hline
		$\rollPitchCost$ &   Roll/pitch error cost     &   \SI{14.6}{\per\radian\squared}    \\ \hline
		$\yawCost$ &   Yaw error cost     &   \SI{3.6}{\per\radian\squared}    \\ \hline
		$\rollPitchRateCost$ &   Roll/pitch rate error cost     &   \SI{0}{\per\radian\squared\second\squared}    \\ \hline
		$\yawRateCost$ &   Yaw rate error cost     &   \SI{0}{\per\radian\squared\second\squared}    \\ \hline
		$\forwardThrustCost$ &   Forward thrust cost     &   \SI{1.0}{\per\newton\squared}    \\ \hline
		$\reverseThrustCost$ &   Reverse thrust cost     &   \SI{2.25}{\per\newton\squared}    \\ \hline
	\end{tabular}
\end{table}

\subsection{Choice of arm angle}
We choose the angle that each arm makes with the diagonal of the vehicle $\armAngle$, as shown in Figure \ref{fig:sysModel}, such that the vehicle is capable of hovering in the two-arms-folded configuration.
That is, the vehicle should be capable of producing a total thrust $\totalThrust$ to offset gravity while producing zero torque on the vehicle.

Let the thrust each propeller can produce be bounded, namely $\thrust{i} \in [\minThrust, \maxThrust]$, where $\minThrust$ and $\maxThrust$ are determined by the physical limits of the powertrain of the vehicle.
Note that in our case $\minThrust < 0$ unlike conventional quadcopters which only allow propellers to spin in the forward direction.

We wish to find $\armAngle$ such that $\twoArmsMixer \thrusts = (m_\Sigma g, 0, 0, 0)$ with $\twoArmsMixer$ as given in \eqref{eq:twoArmsFoldedMixer} while satisfying constraints on the thrusts each propeller can produce.
As the constraints $\tau_x = \tau_y = 0$ can be trivially satisfied for any choice of $\armAngle$ when $\thrust{1} = \thrust{3}$ and $\thrust{2} = \thrust{4}$, we focus on the constraints on the total thrust $\totalThrust$ and yaw torque $\tau_z$:
\begin{gather}
\thrust{1} + \thrust{3} \geq m_\Sigma g \\
-\thrustToTorqueForward\left(\thrust{1} + \thrust{3}\right) - \frac{l\sqrt{2}}{2}\sin\armAngle\left(\thrust{2} + \thrust{4}\right) = 0
\end{gather}

Thus, the following two inequalities must be satisfied in order for the vehicle to be able to hover with two arms folded:
\begin{equation}\label{eq:armAngleIneq}
\begin{gathered}
\armAngle \geq \sin^{-1}\left(\frac{-\thrustToTorqueForward m_\Sigma g}{l\sqrt{2}\minThrust}\right) \\
\maxThrust \geq \frac{m_\Sigma g}{2}
\end{gathered}
\end{equation}

\begin{figure}
	\includegraphics[width=1\columnwidth]{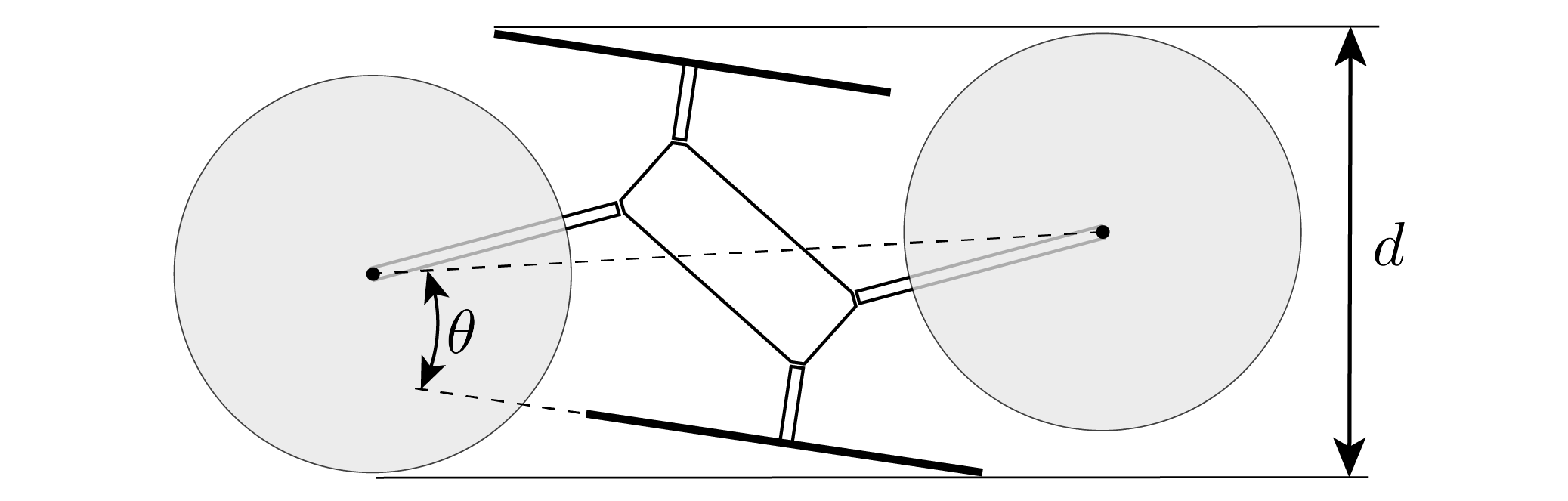}
	\centering
	\caption{Top-down view of the vehicle in the two-arms-folded configuration. The minimum horizontal dimension of the vehicle $d$ increases as the arm angle $\armAngle$ increases.}
	\label{fig:minDimension}
\end{figure}

Because the geometry of the experimental vehicle is defined such that an increase in $\armAngle$ corresponds to an increase in the minimum dimension $\minVehicleDim$ of the vehicle in the two-arms-folded configuration as shown in Figure \ref{fig:minDimension}, we choose the smallest $\armAngle$ such that the vehicle has sufficient control authority to produce reasonable magnitude thrusts and torques with two arms folded.
Specifically, we choose
\begin{equation}\label{eq:armAngleComp}
\armAngle = \sin^{-1}\left(\frac{-\thrustToTorqueForward m_\Sigma g}{l\sqrt{2}\desThrust}\right)
\end{equation}
where $\desThrust > \minThrust$ is the nominal thrust force produced by each of the two folded arms during hover.
Note that if $\desThrust$ is chosen to be very close to $\minThrust$, the vehicle may experience difficulty in compensating for yaw torque disturbances (as the ability of the vehicle to produce yaw torques will be greatly diminished in one direction).
Thus, we choose $\desThrust$ to be roughly half $\minThrust$ (recall $\minThrust < 0$) so that the vehicle is capable of producing roughly equal magnitude yaw torques in each direction.

Note that the bound presented in \eqref{eq:armAngleIneq} is also dependent on several other parameters of the vehicle.
For example, if a smaller $\armAngle$ is desired, it is advantageous to minimize both the mass of the vehicle $m_\Sigma$ and the coefficient $\thrustToTorqueForward$ that relates the thrust produced by each propeller to the torque acting about its rotation axis.
Coincidentally, minimizing $m_\Sigma$ is equivalent to minimizing the power consumption of the vehicle at hover, which is typically a preeminent concern when designing aerial vehicles.

\subsection{Powertrain selection}

As discussed above, the arm angle $\armAngle$ is dependent on both the torque per unit positive thrust produced by each propeller $\thrustToTorqueForward$ as well as the maximum magnitude thrust each propeller can produce when spinning in reverse $\minThrust$.
Thus, in order to minimize $\armAngle$, the ratio between $\thrustToTorqueForward$ and $\minThrust$ must be minimized.
To this end, the powertrain (i.e. battery, speed controllers, motors, and propellers) is chosen such that $\armAngle$ is minimized while simultaneously minimizing the power consumption of the vehicle while flying in the unfolded configuration, as this would likely be the primary mode of operation of the vehicle.
In our model, $\minThrust$ and $\maxThrust$ are determined by the design of the powertrain, and $\thrustToTorqueForward$ and $\thrustToTorqueBack$ are determined by the chosen propellers.

Although we spin several of the propellers in the reverse direction in the two- or four-arms-folded configurations, this does not imply that it would necessarily be advantageous to use symmetric propellers (sometimes referred to as ``3D propellers") which are designed to spin in both directions.
When compared to conventional propellers, symmetric propellers have the advantage of being able to produce much larger thrusts when spinning in reverse (i.e. $\minThrust$ is larger in magnitude), but this comes at the cost of a smaller maximum forward thrust $\maxThrust$ and a larger torque per unit positive thrust $\thrustToTorqueForward$.
Thus, it is possible that the use of symmetric propellers may lead to a larger required $\armAngle$ if the ratio of $\thrustToTorqueForward$ to $\minThrust$ is larger than that of a conventional propeller.
Additionally, $\maxThrust$ must still be large enough to satisfy the constraint given in \eqref{eq:armAngleIneq}, which may be difficult to achieve using symmetric propellers.
Finally, the use of symmetric propellers would greatly increase the power consumption of the vehicle when hovering in the unfolded configuration, as symmetric propellers are not optimized to minimize power consumption compared to conventional propellers.

\begin{figure}
	\includegraphics[width=1\columnwidth]{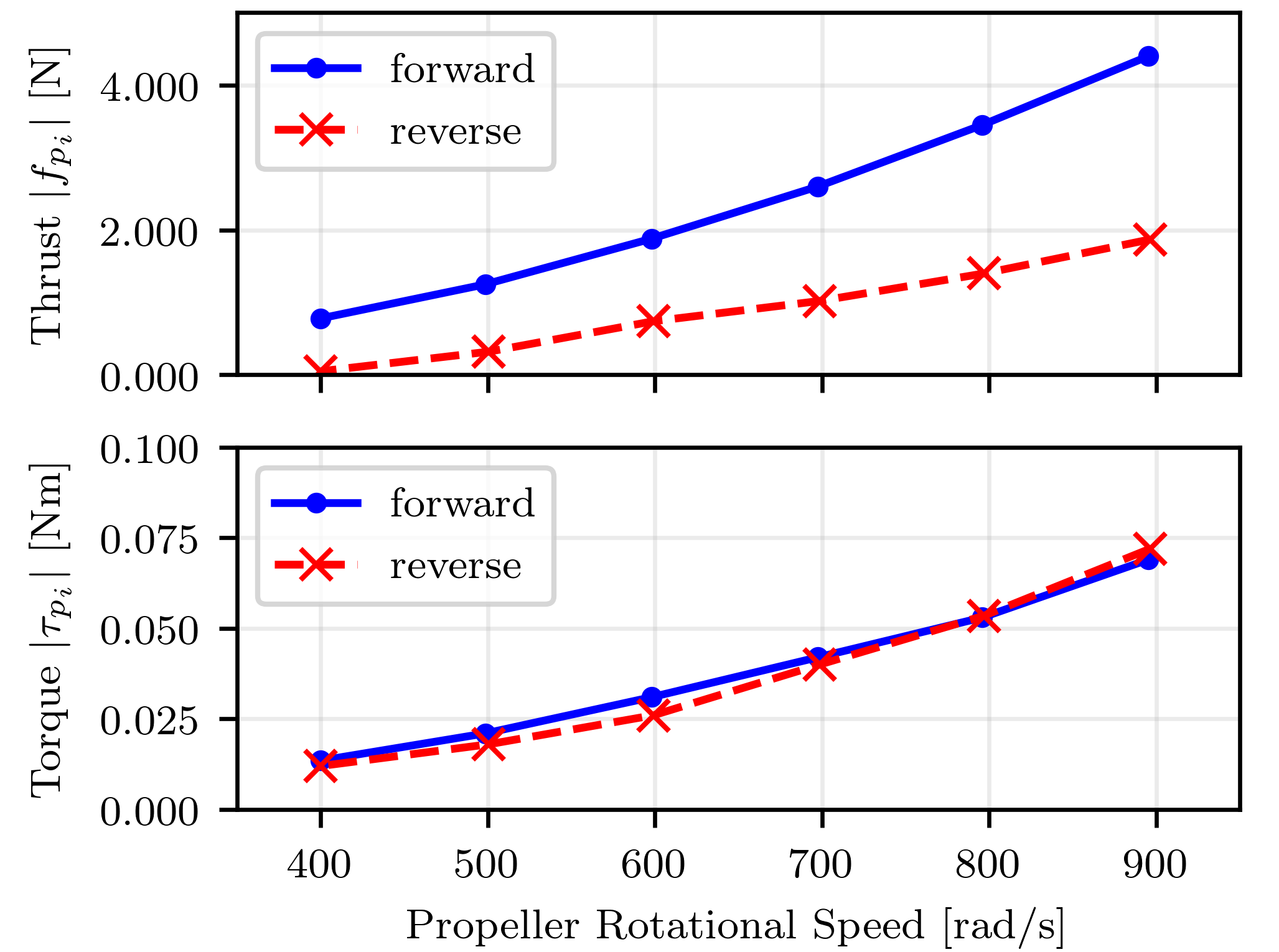}
	\centering
	\caption{Magnitude of thrust and torque produced by an 8038 propeller spinning in both the forward and reverse directions. A load cell capable of measuring forces and torques was used in conjunction with an optical tachometer to collect the data. The propeller produces significantly more thrust but produces roughly the same magnitude torque when spinning in the forward direction compared to the reverse direction for a given speed.}
	\label{fig:propCalib}
\end{figure}

To this end, we choose to use conventional quadcopter propellers on the experimental vehicle.
Figure \ref{fig:propCalib} shows how the thrust and torque produced by a Gemfan 8038 propeller are related to the rotational speed of the propeller, demonstrating the difference in thrust produced by the propeller when spinning in the forward and reverse directions.
We found that the powertrain of the experimental vehicle was capable of driving the propeller to produce up to \SI{3.4}{\newton} of thrust in the reverse direction and 
\SI{7.8}{\newton} of thrust in the forward direction with $\thrustToTorqueForward = \SI{0.0172}{\newton\meter/\newton}$ and $\thrustToTorqueBack = \SI{0.038}{\newton\meter/\newton}$.
This lead to a choice of $\armAngle = \SI{11.9}{\degree}$ according to \eqref{eq:armAngleComp} with $\desThrust = \SI{1.5}{\newton}$.


Finally, we note that although in theory $\thrust{i}$ can achieve any value between $\minThrust$ and $\maxThrust$, in practice we restrict $\thrust{i}$ to not pass through zero unless the vehicle is performing a configuration transition that requires reversing the propeller.
This is due to the fact that we use commonly available electronic speed controllers and brushless motors which use back-EMF to sense the speed of the motor.
The use of back-EMF to sense motor speed results in significantly degraded performance when operating at low speeds (e.g. when changing direction), meaning that such motors are typically restricted to spin in only one direction.
Although this property can affect the performance of the proposed vehicle when changing between configurations, once the propellers have reversed direction they can continue to operate without any significant change in performance.
Thus, in practice we restrict the thrust forces of propellers spinning in the forward direction and reverse direction to be in $[0, \maxThrust]$ and $[\minThrust, 0]$ respectively, and only allow the propellers to change direction when changing between configurations.
We did not specifically design the powertrain of the experimental vehicle to minimize the time required to reverse the propellers, and other powertrain designs (e.g. using ESCs capable of sensing the rotor position) may have improved propeller reversal times.
Such improved reversal times could result in smaller desired state errors when transitioning between configurations, but would likely increase the complexity of the vehicle and incur higher costs.

\subsection{Vehicle Agility}\label{sec:vehicleAgility}

We now examine the effects of the bounds described in Section \ref{sec:thrustLimits} on the experimental vehicle with thrust limits $\minThrust$ and $\maxThrust$.
For notational convenience, we define $\ineqMatrix \in \mathbb{R}^{4\times 4}$ as a matrix with each column defined by $c_{f_i}, c_{x_i}, c_{y_i}$, and $c_{z_i}$ respectively.
Then, the bounds defined in \eqref{eq:thrustLimitPerArm} can be rewritten as:
\begin{equation}\label{eq:foldingLimitIneqs}
\ineqMatrix \begin{bmatrix}
\totalThrust \\
\bodyTorque
\end{bmatrix} \succeq \bs 0
\end{equation}
where $\succeq$ denotes an element-wise inequality, and $\bs 0$ denotes a vector of zeros.

Then, the matrix $\ineqMatrixUnfolded$ for the experimental vehicle in the unfolded configuration is computed to be the following, where the first column has units of meters and the other columns are unitless.
\begin{equation}
\ineqMatrixUnfolded = 
\begin{bmatrix}
0.0144 & -0.0421 & -0.0252 & -1.304 \\
0.0144 & -0.0421 & 0.0252 & 1.304 \\
0.0144 & 0.0421 & 0.0252 & -1.304 \\
0.0144 & 0.0421 & -0.0252 & 1.304 
\end{bmatrix}
\end{equation}

Similarly, the matrix $\ineqMatrixTwoArms$ for the experimental vehicle in the two-arms-folded configuration is computed to be:
\begin{equation}
\ineqMatrixTwoArms = 
\begin{bmatrix}
0.0369 & 0.08 & 0.0225 & 0.0059 \\
0.0237 & 0.345 & -0.289 & 1.26 \\
0.0369 & -0.08 & -0.0225 & 0.0059 \\
0.0237 & -0.345 & 0.289 & 1.26 
\end{bmatrix}
\end{equation}

The individual thrust limits of each propeller can be written in terms of $\totalThrust$ and $\bodyTorque$ by utilizing the inverse of the mapping matrix $M$ introduced in \eqref{eq:mappingDef}:
\begin{equation}\label{eq:thrustLimsAsCtrlInputs}
\begin{bmatrix}
\bs I \\ - \bs I
\end{bmatrix}
M^{-1}
\begin{bmatrix}
\totalThrust \\ \bodyTorque
\end{bmatrix} 
\succeq 
\begin{bmatrix}
\bs 1 \minThrust \\
-\bs 1 \maxThrust
\end{bmatrix}
\end{equation}
where $\bs I$ the $4 \times 4$ identity matrix, and $\bs 1$ is vector of ones of length four.

In order to compare the agility of the experimental vehicle to a conventional quadcopter, we examine how the set of feasible values of $\totalThrust$ and $\bodyTorque$ is reduced when imposing the bounds given in \eqref{eq:foldingLimitIneqs} (i.e. those that prevent the arms from folding or unfolding).
Note that both the experimental vehicle and a conventional quadcopter must satisfy the bounds on $\totalThrust$ and $\bodyTorque$ given by \eqref{eq:thrustLimsAsCtrlInputs} (i.e. those that ensure $\thrust{i} \in [\minThrust, \maxThrust]$), but that the experimental vehicle must additionally satisfy the bounds that prevent the arms from folding or unfolding.

\begin{figure}
	\includegraphics[width=1\columnwidth]{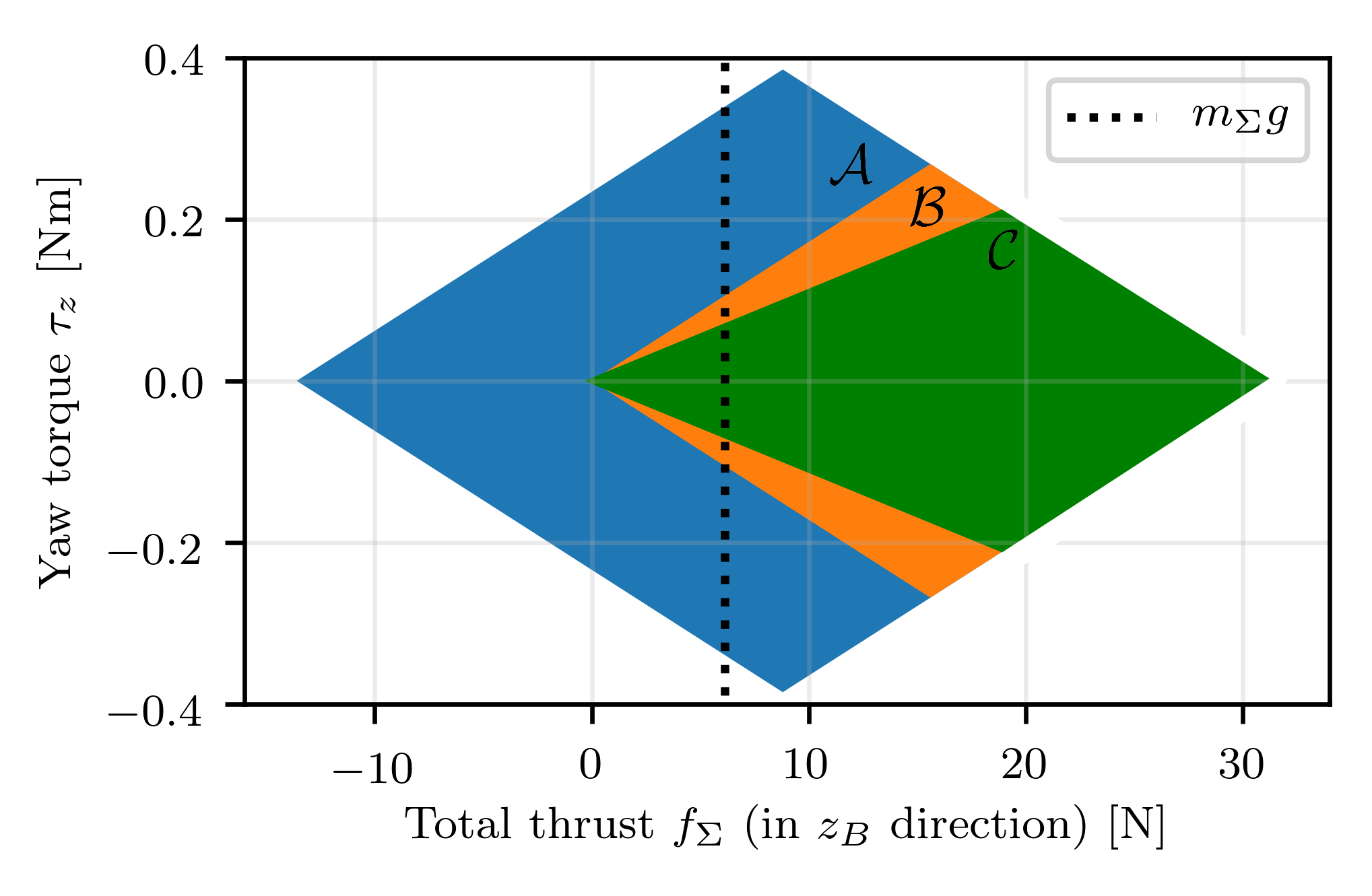}
	\centering
	\caption{Range of feasible total thrusts $\totalThrust$ and yaw torques $\tau_z$ for the experimental vehicle in the unfolded configuration with zero roll and pitch torques $\tau_x = \tau_y = 0$. The dotted black line denotes the value of $\totalThrust$ at hover. The blue set $\mathcal{A}$ represents the feasible inputs when only the constraints on the minimum and maximum thrusts of each propeller $\minThrust$ and $\maxThrust$ are considered. The orange set $\mathcal{B}$ represents the feasible inputs for a conventional quadcopter, i.e. with $\minThrust = 0$ rather than $\minThrust < 0$. Finally, the green set $\mathcal{C}$ represents the feasible inputs when the constraints that prevent the arms from folding are imposed, primarily reducing the range of feasible yaw torques. Note that $\mathcal{C} \subset \mathcal{B} \subset \mathcal{A}$.}
	\label{fig:feasInputs}
\end{figure}

The reduction in agility of the experimental vehicle when $\tau_x = \tau_y = 0$ is shown in Figure \ref{fig:feasInputs}, where we observe how the set of feasible yaw torques $\tau_z$ and total thrusts $\totalThrust$ is reduced in comparison to a conventional quadcopter.
As shown in the figure, the bounds that prevent the arms from folding primarily result in a reduction in the range of feasible yaw torques.
Specifically, the maximum yaw torque the experimental vehicle can produce at hover (i.e. when $\totalThrust = m_\Sigma g$ and $\tau_x = \tau_y = 0$) is reduced by 36\% when compared to a conventional quadcopter.
In terms of agility, this means that when hovering, the experimental vehicle would take a minimum of \SI{1.2}{\second} to rotate \SI{180}{\degree} in yaw (when starting and ending at rest), whereas a conventional quadcopter would take only \SI{0.96}{\second}.
Compared to our previous work \cite{bucki2019design} which used springs to fold the arms, the experimental vehicle can produce roughly 2.8 times higher yaw torques at hover.

A similar analysis of the maximum magnitude roll and pitch torques the vehicle can produce at hover shows them to be no less than those of a conventional quadcopter, indicating that the bounds that prevent the arms from folding given in \eqref{eq:foldingLimitIneqs} are actually less restrictive than those on each of the individual thrust forces given in \eqref{eq:thrustLimsAsCtrlInputs}.
Finally, we find that the minimum and maximum total thrust forces are also no less than those of a conventional quadcopter, which is also an improved result from our previous work \cite{bucki2019design} where we found that the minimum total thrust force was 70\% of the thrust force required to hover (again due to the use of springs to fold the arms).
Thus, this analysis implies that the only significant tradeoff between the proposed vehicle design and a conventional quadcopter (in terms of the control authority of the vehicle) is the reduction of the maximum yaw torque the vehicle can produce.



%% file: experiment.tex
\section{Experimental results} \label{sec:experiments}

In this section, we show how the agility of the proposed vehicle compares to that of a conventional quadcopter, and demonstrate how the ability of the proposed vehicle to fold and unfold each arm enables it to perform a number of tasks which would be difficult or impossible to perform using a conventional quadcopter.

We first examine how the tightness of the control input bounds (used to prevent the arms from folding) change with time when flying a figure eight trajectory in the unfolded configuration.
This demonstrates how the agility of the proposed vehicle compares to a conventional quadcopter, specifically the reduced ability of the proposed vehicle to produce yaw torques.
Next we demonstrate the ability of the vehicle to traverse narrow horizontal tunnels and vertical gaps by transitioning to the two-arms-folded and four-arms-folded configurations respectively.
We then show how the vehicle can perch on wires by folding all four arms, and perform limited grasping tasks in the two-arms-folded configuration.
\footnote{Videos of each of the experiments discussed in this section can be viewed in the attached video or at \url{https://youtu.be/1GUyAQxVEtg}}

\subsection{Flight in Unfolded Configuration}\label{sec:figEight}

\begin{figure}
	\includegraphics[width=\columnwidth]{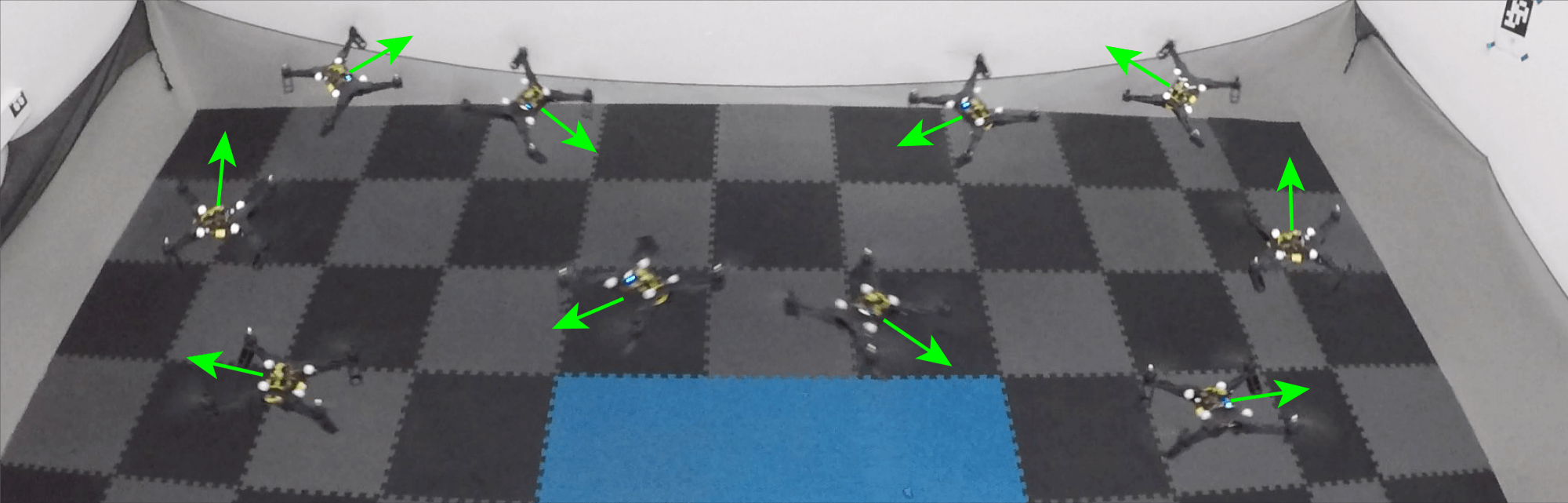}
	\centering
	\caption{Composite image of the vehicle flying a figure eight trajectory in the unfolded configuration. Green arrows show the heading of the vehicle at each point.}
	\label{fig:figEightExperiment}
\end{figure}

We first demonstrate the ability of the proposed vehicle to fly in the unfolded configuration, i.e. as a conventional quadcopter.
The vehicle was commanded to fly in a figure eight trajectory, where the yaw angle of the vehicle is constrained such that the vehicle faces along the trajectory.
Figure~\ref{fig:figEightExperiment} shows the trajectory of the vehicle during the experiment.
Specifically, the trajectory is defined as
\begin{gather}\label{eq:figEight}
	x = 0.75\sin \roundB{\omega t}, \qquad y = 1.5\sin 
	\roundB{\frac{\omega}{2} t}, \qquad z = 1.5 \\
	\psi = \textrm{atan2}\roundB{\dot{y}, \dot{x}} \nonumber
\end{gather}
where $x$, $y$, and $z$ are the desired position of the center of mass of the vehicle in meters, $\psi$ is the desired yaw angle, and $\omega$ governs the speed of the vehicle along the figure eight trajectory.

\begin{figure}
	\includegraphics[width=1\columnwidth]{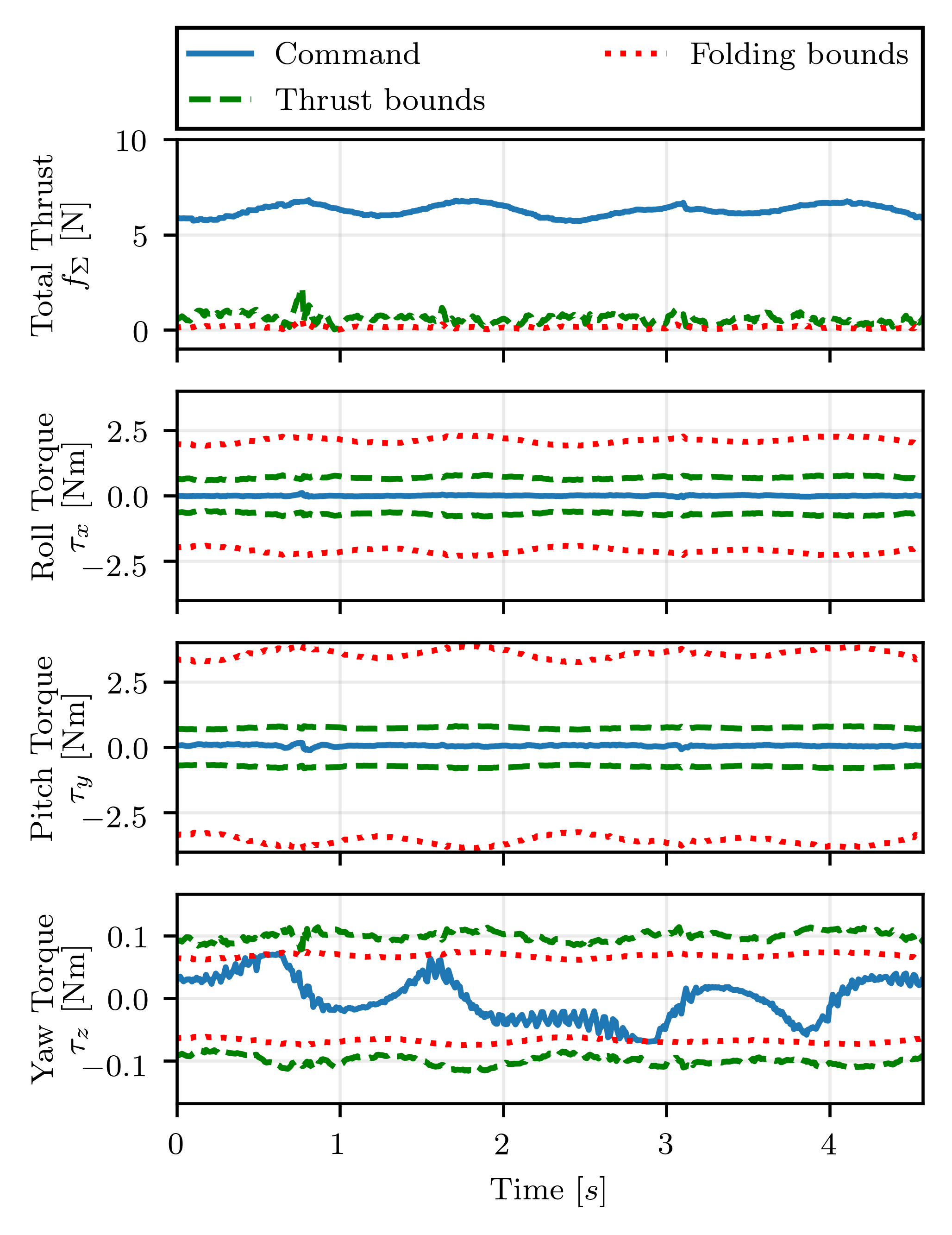}
	\centering
	\caption{Total thrust and torque commands (drawn as solid blue lines) applied while flying one cycle of a figure eight trajectory with $\omega = \SI{2.75}{\radian\per\second}$ in the unfolded configuration.
		The dashed green lines show the bounds on the commands due to the minimum and maximum thrust each propeller can produce (i.e. the bounds used for a conventional quadcopter), while the dotted red lines show the bounds used to prevent the arms from folding (i.e. the additional bounds placed on the proposed vehicle).
		As shown, the folding bounds are only more restrictive than the thrust bounds for $\tau_z$.
		Note that the folding bounds impose no upper limit on the total thrust, and thus for clarity only the lower bounds on the total thrust are shown.}
	\label{fig:fig8}
\end{figure}

Figure~\ref{fig:fig8} shows how the commanded total thrust $\totalThrust$ and roll, pitch, and yaw torques ($\tau_x$, $\tau_y$, and $\tau_z$) change over one cycle of the figure eight trajectory with $\omega = \SI{2.75}{\radian\per\second}$.
The choice of $\omega = \SI{2.75}{\radian\per\second}$ results in a trajectory where the bounds that prevent the arms from folding become active, but the bounds on the minimum and maximum thrust forces produced by each propeller do not (i.e. where a conventional quadcopter does not hit its input bounds, but the proposed vehicle does).
Two different bounds on the commands are shown, representing (a) those imposed by the minimum/maximum thrust each individual propeller can produce (i.e. \eqref{eq:thrustLimsAsCtrlInputs}), shown as dashed green lines, and (b) those imposed to prevent the arms from folding (i.e. \eqref{eq:foldingLimitIneqs}), shown as dotted red lines.
Both sets of bounds are computed assuming the hierarchical reduction of commands described in Section \ref{sec:thrustLimits} is used, meaning that if a given bound is not satisfied, $\tau_z$ is reduced until the bound is satisfied or $\tau_z = 0$.
Thus, the bounds on $\totalThrust$, $\tau_x$, and $\tau_y$ are computed assuming that $\tau_z$ can be reduced to zero, while the bounds on $\tau_z$ are computed assuming that $\totalThrust$, $\tau_x$, and $\tau_y$ remain unchanged.

While a conventional quadcopter is only constrained by the thrust bounds, the proposed vehicle is constrained by both the thrust bounds and the folding bounds.
For this specific trajectory, the minimum value of $\totalThrust$ that satisfies the folding bounds remains  below or equal to the value given by the thrust bounds.
This implies that for the entire figure eight trajectory, the bounds on the total thrust of the proposed vehicle are identical to those imposed on a conventional quadcopter (note the folding bounds do not impose an upper bound on $\totalThrust$).
Similarly, for $\tau_x$ and $\tau_y$, the folding bounds are a magnitude less restrictive than the thrust bounds, implying that the proposed vehicle is equally capable of producing roll and pitch torques throughout the experiment as a conventional quadcopter.
However, the folding bounds are more restrictive than the thrust bounds for $\tau_z$, meaning that the proposed vehicle cannot track as aggressive yaw trajectories as a conventional quadcopter.

Because the proposed vehicle does not have significantly different geometry, components, or inertial properties than a conventional quadcopter, its tracking performance will only differ from that of a conventional quadcopter when the folding bounds on $\totalThrust$, $\tau_x$, $\tau_y$, or $\tau_z$ are active.
In order to demonstrate this, the vehicle was commanded to fly ten cycles of \eqref{eq:figEight} with $\omega = \SI{2.25}{\radian\per\second}$ and again with $\omega = \SI{2.75}{\radian\per\second}$.
For $\omega = \SI{2.25}{\radian\per\second}$, the folding bounds were not violated, whereas for $\omega = \SI{2.75}{\radian\per\second}$ the commanded yaw torque $\tau_z$ was high enough to be constrained by the folding bounds at several points, as shown in Figure~\ref{fig:fig8}.
The thrust bounds were not violated for either trajectory.

In order to compare the tracking performance of the vehicle to a conventional quadcopter, the hinges of the experimental vehicle were tightened such that the arms could not fold due to friction, and the same trajectories were flown without enforcing the folding bounds.
With $\omega = \SI{2.25}{\radian\per\second}$, the average position and yaw tracking errors were \SI{16.6}{\centi\meter} (\SI{24.6}{\centi\meter} max) and \SI{8.7}{\degree} (\SI{23.2}{\degree} max) for the vehicle with folding arms, and were \SI{16.7}{\centi\meter} (\SI{25.3}{\centi\meter} max) and \SI{9.1}{\degree} (\SI{23.5}{\degree} max) for the conventional quadcopter.
With $\omega = \SI{2.75}{\radian\per\second}$, the average position and yaw tracking errors were \SI{19.2}{\centi\meter} (\SI{27.7}{\centi\meter} max) and \SI{10.3}{\degree} (\SI{29.4}{\degree} max) for the vehicle with folding arms, and were \SI{19.1}{\centi\meter} (\SI{28.2}{\centi\meter} max) and \SI{10.0}{\degree} (\SI{26.9}{\degree} max) for the conventional quadcopter.

This aligns with our previous analysis as described in Section \ref{sec:vehicleAgility}.
That is, we could not show any significant difference in the position tracking capabilities of the proposed vehicle compared to a conventional quadcopter, but the proposed vehicle did experience larger maximum yaw tracking errors than a conventional quadcopter (in this case \SI{29.4}{\degree} versus \SI{26.9}{\degree} with $\omega = \SI{2.75}{\radian\per\second}$).
However, these differences in yaw tracking error are still relatively small, as the figure eight trajectory only results in $\tau_z$ violating the folding bound for very brief periods of time.
A more significant reduction in yaw tracking performance would be expected if the vehicle is required to produce large yaw torques for long periods of time (e.g. if flying in an environment with large yaw torque disturbances, or when rapidly changing the desired yaw angle as mentioned in Section \ref{sec:vehicleAgility}).

\subsection{Horizontal flight through a narrow tunnel}\label{sec:tunnel}

\begin{figure}
	\includegraphics[width=\columnwidth]{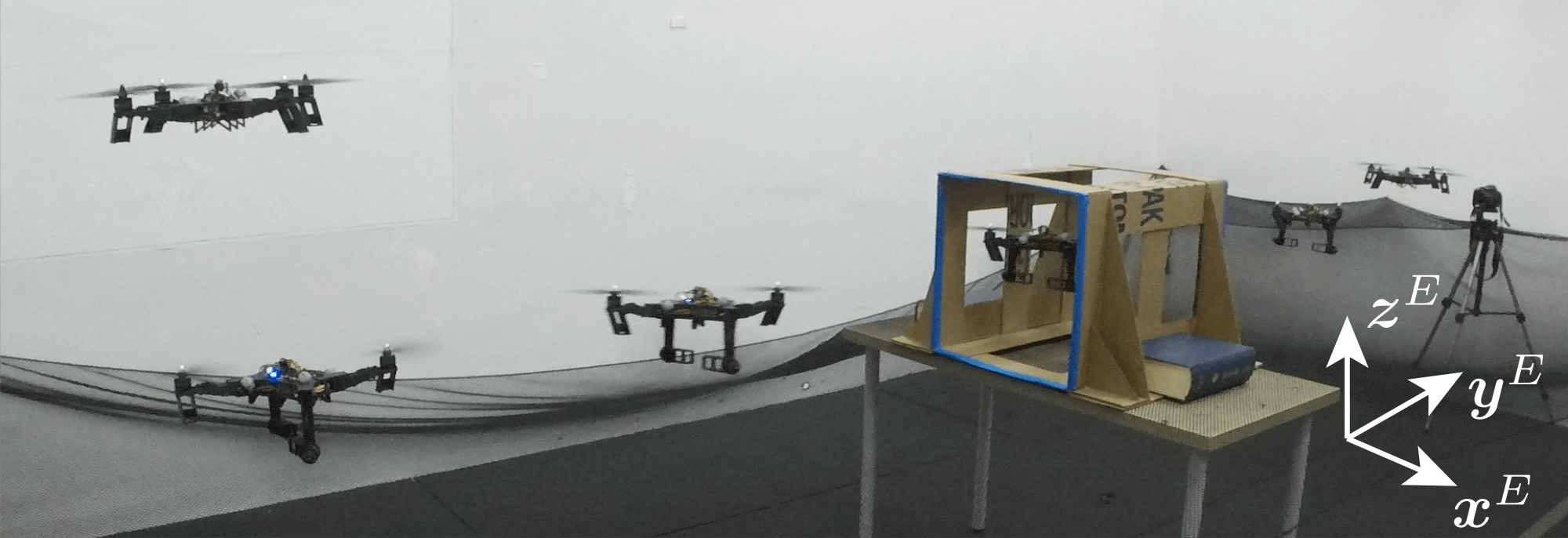}
	\centering
	\caption{Composite image of the vehicle transitioning from the unfolded to two-arms-folded configuration (left), flying through a narrow tunnel, and transitioning back to the unfolded configuration (right).}
	\label{fig:tunnelExperiment}
\end{figure}

\begin{figure}
	\includegraphics[width=1\columnwidth]{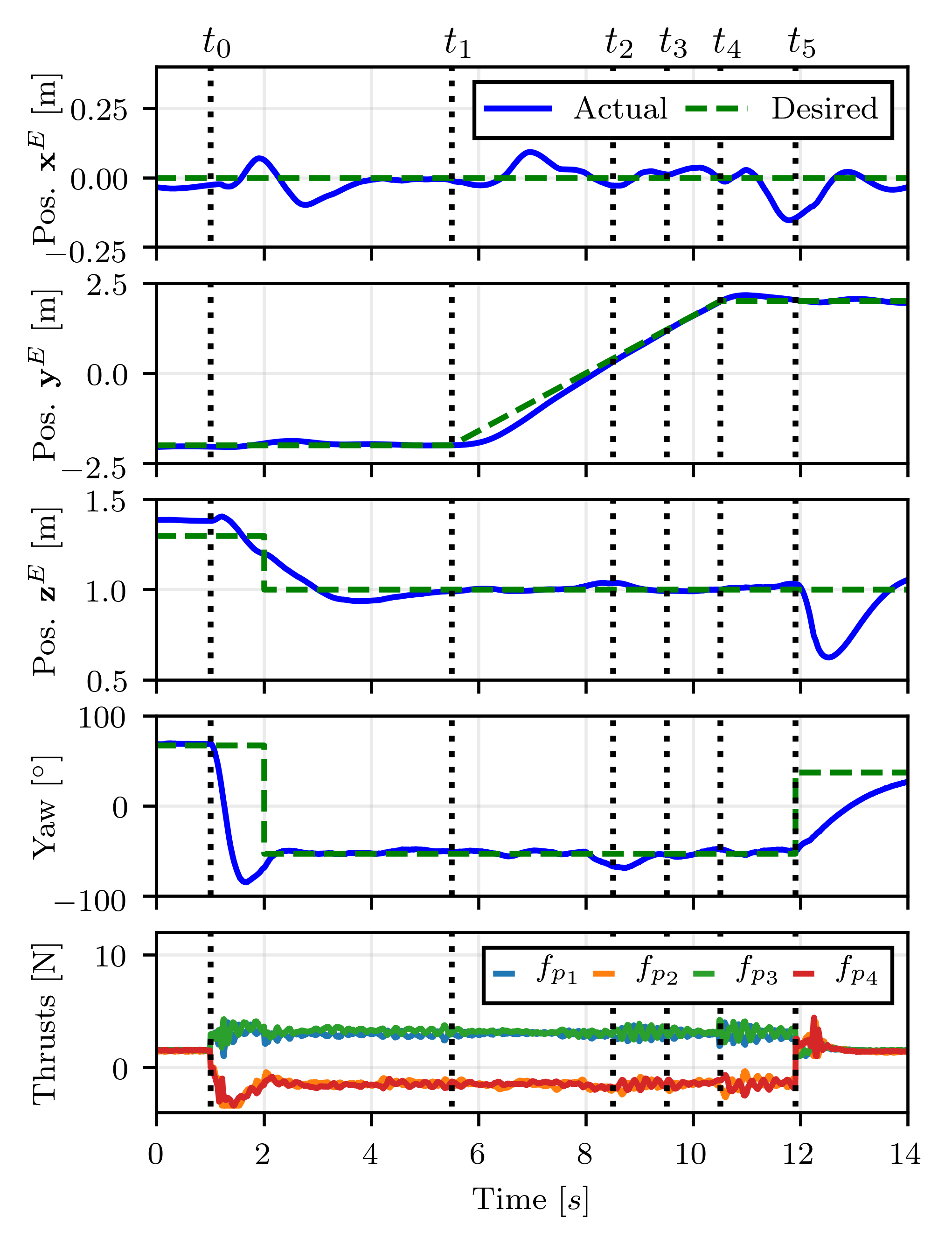}
	\centering
	\caption{Trajectory of the vehicle while traversing a narrow tunnel. The vehicle starts to transition from the unfolded to the two-arm-folded configuration at time $t_0$, changing its yaw angle and lowering its height during the transition. At time $t_1$ the vehicle begins moving towards the tunnel, and enters the tunnel at time $t_2$. The vehicle exits the tunnel at time $t_3$, returns to hover at time $t_4$, and finally transitions back to the unfolded configuration at time $t_5$.}
	\label{fig:tunnelTraj}
\end{figure}

\begin{figure}
	\includegraphics[width=1\columnwidth]{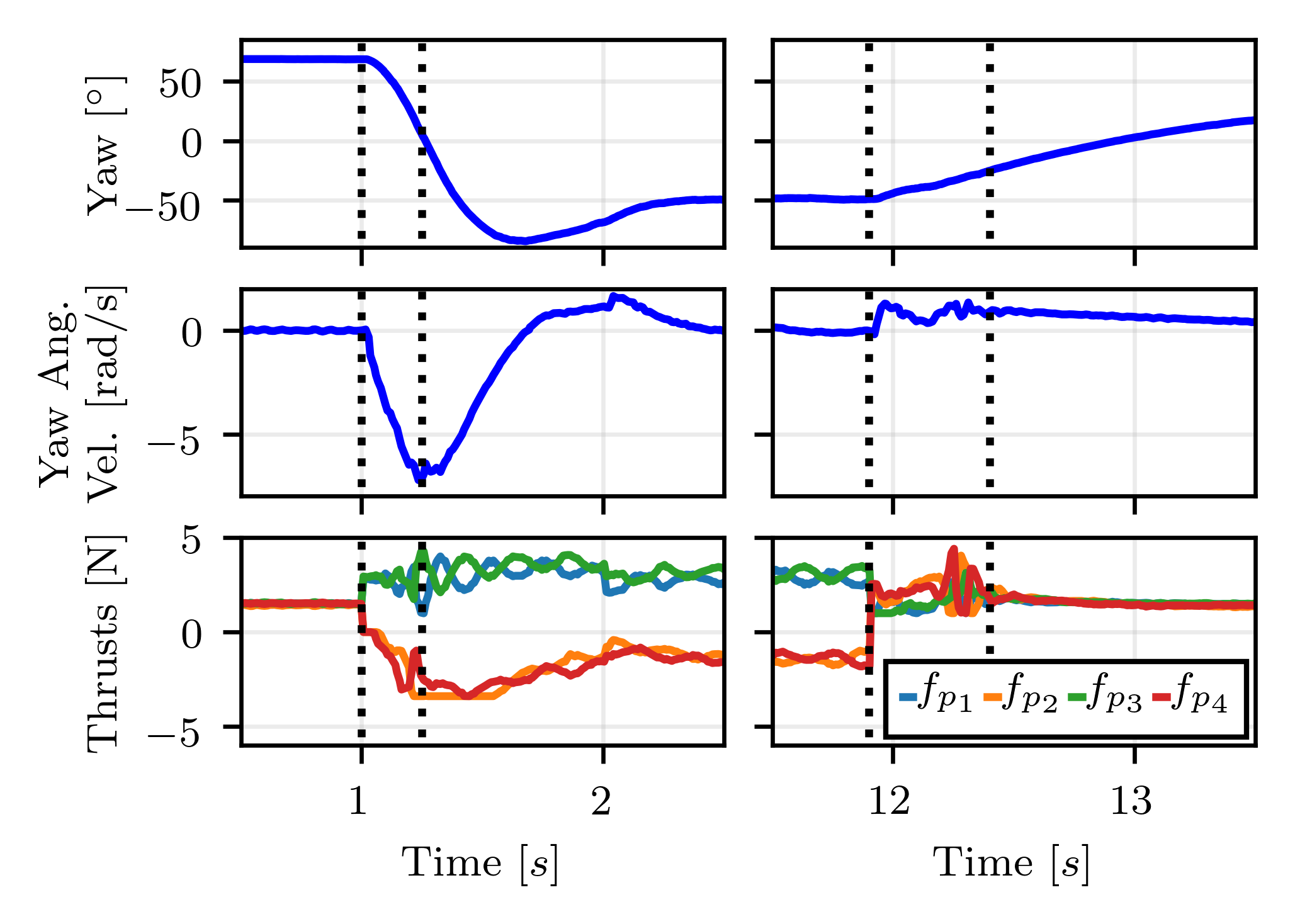}
	\centering
	\caption{Transition from the unfolded to the two-arms-folded configuration (left) and back (right). The black dotted lines denote when arms 2 and 4 begin and finish transitioning between the folded and unfolded configurations. When transitioning to the two-arms-folded configuration, the vehicle experiences a large change in yaw angular velocity due to the yaw torque produced by propellers 1 and 3 while propellers 2 and 4 are changing direction. In contrast, a much smaller change in yaw angular velocity is experienced when transitioning to the unfolded configuration, as the yaw torque produced by arms 1 and 3 during the transition is much less.}
	\label{fig:transitions}
\end{figure}

We now demonstrate how the proposed vehicle can be used to fly in confined spaces which would normally be inaccessible to a conventional quadcopter of similar size, and provide a brief analysis of the configuration transitions required to do so.
The vehicle was flown through a tunnel with a cross section that measures \SI{43}{\centi\meter} by \SI{43}{\centi\meter}, as shown in Figure~\ref{subfig:tunnel}.
These dimensions were chosen such that the vehicle could not traverse the tunnel in the unfolded configuration even with perfect trajectory tracking, as the minimum width of the vehicle in the unfolded configuration is \SI{43}{\centi\meter}.
However, the minimum width of the vehicle in the two-arms-folded configuration is \SI{24}{\centi\meter}, allowing it to pass even with imperfect tracking.

To perform the maneuver, shown in Figures \ref{fig:tunnelExperiment} and \ref{fig:tunnelTraj}, the vehicle first transitions from the unfolded configuration to two-arms-folded configuration, then flies through the tunnel, and finally transitions back to the unfolded configuration.
The yaw angle of the vehicle was chosen to maximize the distance of the vehicle from the walls of the tunnel when flying through its center.
When entering the tunnel, the proposed vehicle experiences aerodynamic forces due to interaction between the walls of the tunnel and the airflow produced by the folded arms, producing a small yaw error that is quickly compensated for by the attitude controller at time $t_2$ as shown in Figure~\ref{fig:tunnelTraj}.
This ``wall effect" is similar to ground effect, which has been analyzed with respect to quadcopter in e.g. \cite{bernard2017dynamic}.
Although reduced tracking errors could be achieved by modeling and compensating for these effects, in practice we have found our controller to be sufficiently capable of compensating for ``wall effect" such that an acceptable level of performance can be achieved (i.e. a \SI{43}{\centi\meter} by \SI{43}{\centi\meter} tunnel can be reliably traversed).

Figure~\ref{fig:transitions} shows a more detailed view of the vehicle transitioning between the unfolded and two-arms-folded configurations before traversing the tunnel, and its transition back to the unfolded configuration afterwards.
The arms take \SI{0.25}{\second} to fold and \SI{0.5}{\second} to unfold during the transitions.
This difference can be partially explained by the fact that while arms 1 and 3 support the vehicle during the transitions, a gravitational force acts on arms 2 and 4 that makes it easier to fold the arms than unfold the arms.
As discussed in Section \ref{sec:transitions}, the vehicle ends the transition to the two-arms-folded configuration with a large yaw angular velocity (approximately \SI{7}{\radian\per\second}) due to the yaw torques applied by the upward facing propellers (1 and 3) while propellers 2 and 4 reverse directions, as well as the change in angular momentum of the propellers.
However, a much smaller change in yaw angular velocity occurs when transitioning back to the unfolded configuration due to the fact that the upward facing propellers produce comparatively less yaw torque.
This difference arises from the fact that when flying in the two-arms-folded configuration the upward facing propellers must produce roughly twice as much thrust (and thus twice as much yaw torque) to support the vehicle than when flying in the unfolded configuration.

As discussed in Section \ref{sec:relWork}, several other quadcopter-like designs exist which allow the vehicle to change shape in order to traverse narrow tunnels.
However, all such designs require one or more actuators to be added beyond those required by a conventional quadcopter, increasing the mass of the vehicle (and thus decreasing flight time) compared to both a conventional quadcopter and the proposed vehicle when flying in the unfolded configuration.
In contrast, due to the high propeller speeds required to hover in the two-arms-folded configuration, the herein proposed vehicle likely requires significantly more power than other morphing aerial vehicles to traverse narrow spaces.
Thus, the proposed vehicle is likely most useful (compared to other morphing vehicles) in situations requiring long flight times in the unfolded configuration with only periodic transitions into and out of the two-arms-folded configuration.

\subsection{Vertical flight through a narrow gap}\label{sec:vertGap}

\begin{figure*}
	\includegraphics[width=\textwidth]{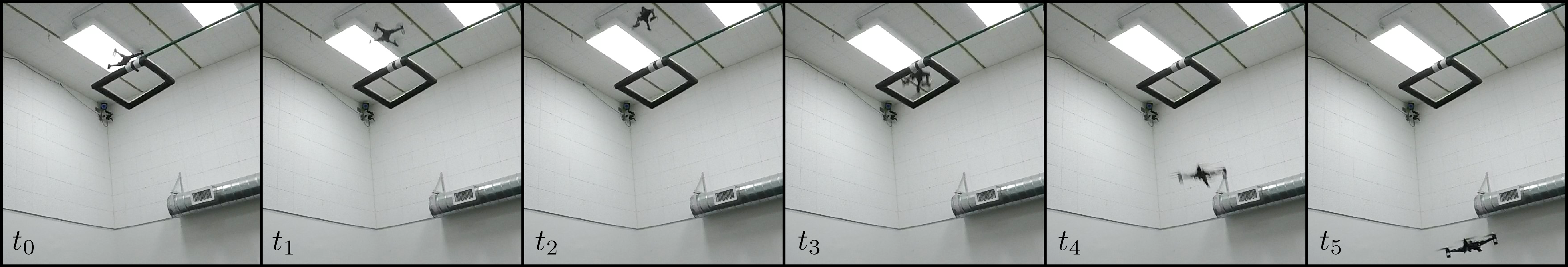}
	\centering
	\caption{Image sequence of the vehicle transitioning from the unfolded to the four-arms-folded configuration and back in order to traverse a narrow gap. Data associated with this experiment is shown in Figure~\ref{fig:gapTrajData}.}
	\label{fig:gapExperiment}
\end{figure*}

\begin{figure}
	\includegraphics[width=1\columnwidth]{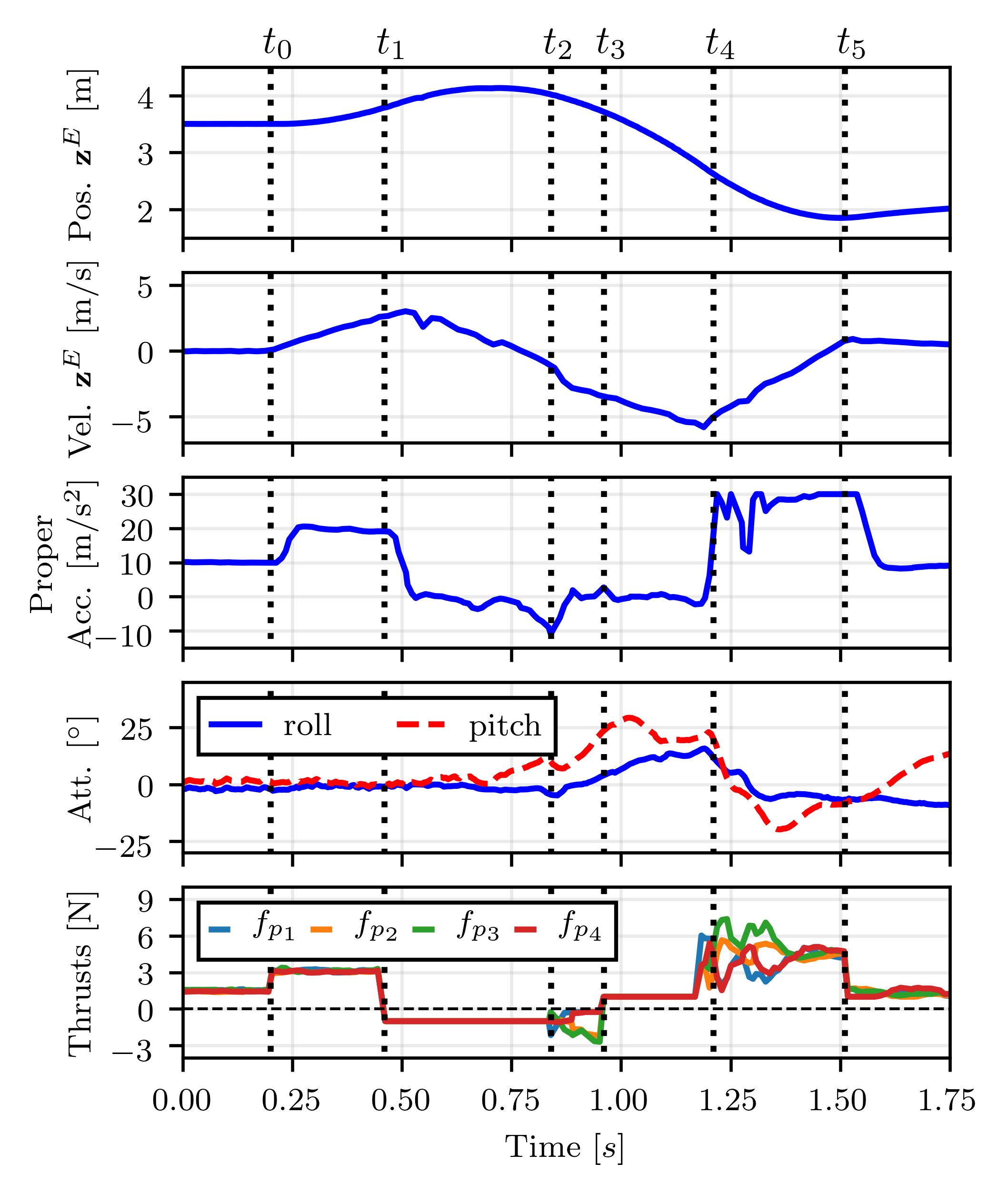}
	\centering
	\caption{Trajectory of the vehicle while passing downward through a narrow gap in the four-arms-folded configuration. The position and velocity of the vehicle are given in the vertical \z{E}{} direction as measured by the motion capture system, and the proper acceleration is given in the \z{B}{} direction as measured by the onboard accelerometer. The vehicle starts accelerating upward at time $t_0$, and commands each propeller to produce a constant negative thrust at time $t_1$, initiating the transition to the four-arms-folded configuration. At time $t_2$ the arms finish folding, and the four-arms-folded controller is used to stabilize the attitude of the vehicle. Next, at time $t_3$, a constant positive thrust command is sent to each motor to initiate the transition back to the unfolded configuration, resulting in the vehicle returning to the unfolded configuration at time $t_4$. Finally, the vehicle is commanded to accelerate upward to reduce its downward velocity until the vehicle comes to rest at time $t_5$.}
	\label{fig:gapTrajData}
\end{figure}

Next, we demonstrate capability of the vehicle to fold all four arms during flight, allowing for passage through narrow gaps in projectile motion.
The maneuver is inspired in part by how birds fold their wings when passing through narrow gaps, as shown in \cite{schiffner2014minding}, and mirrors our previous work \cite{bucki2019design}, where we demonstrated a similar capability using springs to fold the arms rather than reverse thrust forces.
Here we only show the vehicle traversing a gap vertically, as the traversal of gaps in the horizontal direction can be more easily accomplished using the two-arms folded method demonstrated in Section \ref{sec:tunnel}.
The gap measures \SI{43}{\centi\meter} by \SI{43}{\centi\meter}, and the experimental vehicle measures \SI{27}{\centi\meter} by \SI{35}{\centi\meter} in the four-arms-folded configuration.

Figure~\ref{fig:gapExperiment} shows images of the gap traversal maneuver, and Figure~\ref{fig:gapTrajData} graphs the trajectory of the vehicle during the maneuver, which consists of the following stages.
First, the vehicle aligns itself with the gap while hovering above it.
Once aligned, the vehicle begins to accelerate upward from time $t_0 = \SI{0.2}{\second}$ to time $t_1 = \SI{0.46}{\second}$.
After completing this upward trajectory, a constant thrust command of \SI{-1}{\newton} is sent to each propeller at time $t_1$.
At time $t_2 = \SI{0.84}{\second}$ the arms finish the transition to the folded configuration, and the four-arms-folded attitude controller is used to stabilize the vehicle, where the desired attitude is chosen such that \z{B}{} is in the vertical direction.
Next, at time $t_3 = \SI{0.96}{\second}$, a constant thrust command of \SI{1}{\newton} is sent to each propeller in order to unfold the arms.
The vehicle traverses the gap (located at \SI{3.3}{\meter} in this experiment) at approximately this time.
Then, at time $t_4 = \SI{1.21}{\second}$, the arms finish unfolding as evidenced by a sharp increase in the acceleration of the vehicle in the \z{B}{} direction.
At this time the unfolded configuration controller is once again enabled, and the vehicle is commanded to produce a large vertical acceleration until the vertical speed of the vehicle is reduced to zero, which occurs at time $t_5 = \SI{1.51}{\second}$.

Note that although using larger constant thrust commands than \SI{1}{\newton} to fold and unfold the arms would result in the arms folding/unfolding more quickly, in practice we have found it preferable to command smaller constant thrust values.
This is due to the fact that the arms may not fold at exactly the same time (e.g. due to friction), and thus large constant thrusts may result in large torques being exerted on the vehicle, leading to potentially large attitude errors once the transition is completed.
The reduction of attitude errors in the four-arms-folded configuration is crucial because it ensures that the thrust direction of the vehicle will be in the opposite direction of its velocity after transitioning back to the unfolded configuration, allowing for the vehicle to quickly recover to a hover state.


\subsection{Wire perching}
The vehicle is also capable of perching on wires in the four-arms-folded configuration, as shown in Figure~\ref{subfig:perch}.
To perform this maneuver, the vehicle simply aligns itself with the wire and lands on top of it, turning off all four motors when the maneuver is complete.
The body of the experimental vehicle includes a notch that runs the length of the central body the vehicle, which helps align the vehicle with the wire when perching.
Because only the central body of the vehicle is supported by the wire, the four arms fold downward.
This shifts the center of mass of the vehicle below the wire, which allows the vehicle to perch on the wire in a stable configuration.
For the experimental vehicle, the center of mass is shifted \SI{4}{\centi\meter} downward by folding the arms, resulting in the center of mass of the vehicle being \SI{2}{\centi\meter} below where the wire contacts the vehicle.

This method differs from existing perching methods (e.g. \cite{popek2018autonomous}, \cite{hang2019perching}, \cite{doyle2012avian}) in that it does not require the vehicle to carry an additional gripper to perform the perching task, allowing for the proposed vehicle to be lighter than vehicles of a similar size which require gripper mechanisms.
However, because the proposed vehicle does not rely on grippers or adhesives to attach to the wire when perching, it may have difficulty in remaining attached to the wire when large disturbances are present (e.g. in high winds or when the wire is swaying).



\subsection{Grasping}
\begin{figure}
	\includegraphics[width=\columnwidth]{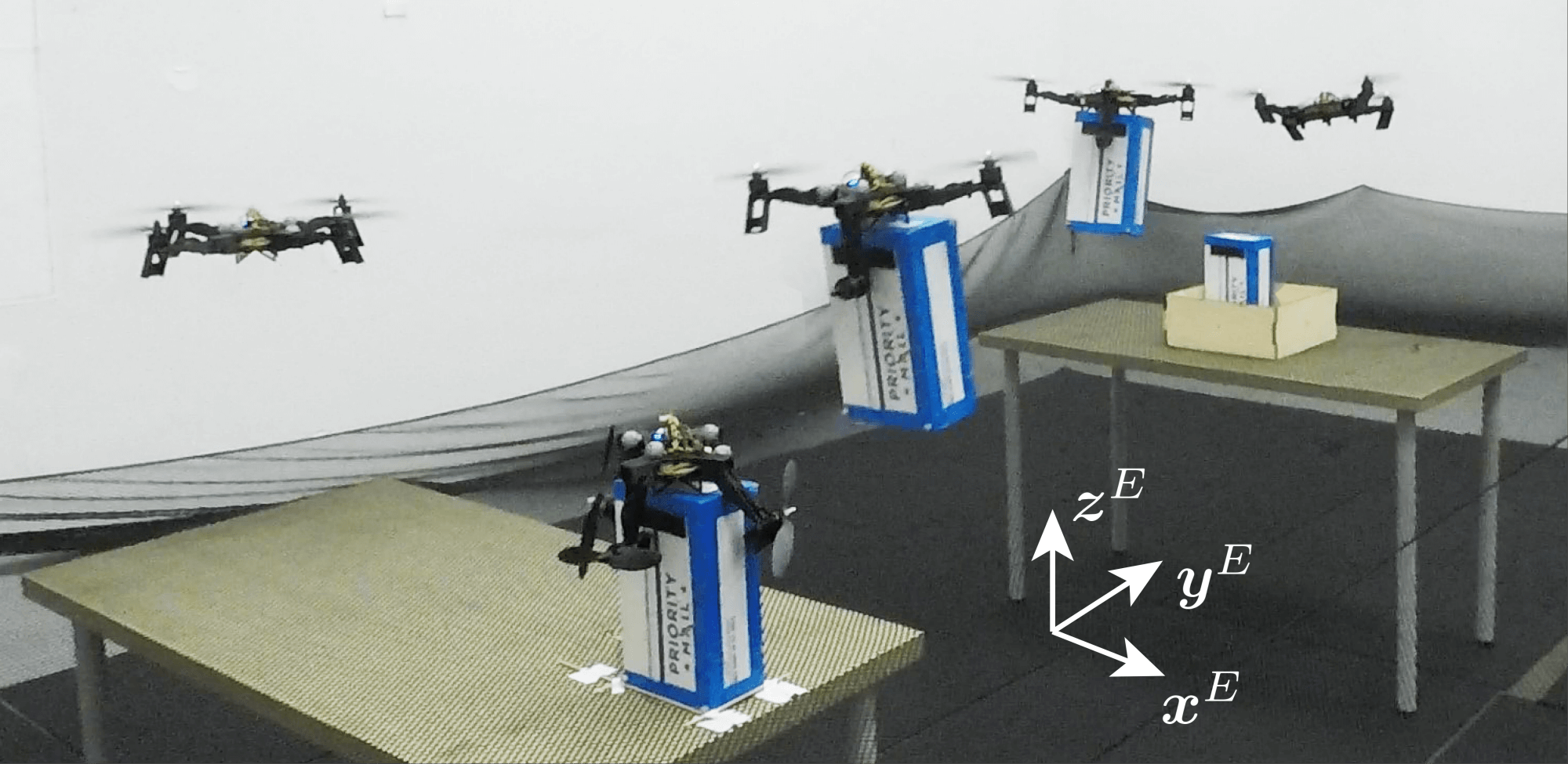}
	\centering
	\caption{Composite image of the vehicle grasping a box (left), flying it to a new location, and dropping the box by returning to the unfolded configuration (right).}
	\label{fig:boxExperiment}
\end{figure}

Finally, we show how the two-arms-folded configuration can be used to perform a simple grasping task, as shown in Figure~\ref{fig:boxExperiment}.
In this experiment a box with a mass of \SI{83}{\gram} that measures $\SI{9}{\centi\meter} \times \SI{15}{\centi\meter} \times \SI{25}{\centi\meter}$ in height is used.
The box was specifically chosen to be \SI{9}{\centi\meter} in width in order to  allow for the box to be grasped without significantly changing the geometry of the two-arms-folded configuration, as the distance between the legs of two opposing folded arms is approximately \SI{9}{\centi\meter}.
Note that because the total mass of the vehicle $m_\Sigma$ increases when holding the box, each of the bounds given in \eqref{eq:armAngleIneq} that govern the ability of the vehicle to hover in the two-arms-folded configuration become more restrictive, significantly limiting the maximum mass of a box that can be carried.


The experiment was conducted as follows:
The vehicle was first commanded to land on top of the box, which was constrained such that it could not rotate in the yaw direction.
After landing, all four propellers were disabled, allowing two of the arms of the vehicle to fall into grasping position.
Next, the two arms used to grasp the box were commanded to produce a thrust of \SI{-2}{\newton} for one second to allow the arms to settle into a firm grasping position, after which time the two unfolded arms were commanded to produce a small thrust of \SI{1}{\newton} for one second such that they fully unfolded before takeoff.
After this grasping procedure was completed, the vehicle was commanded to takeoff and fly to the desired drop-off location using the two-arms-folded configuration controller, which was modified to account for the change in location of the center of mass of the vehicle as discussed in Section \ref{sec:indivThrustComp}.
After flying to the drop-off location, the vehicle was commanded to transition back to the unfolded configuration, resulting the box being released at the desired location.

\begin{figure}
	\includegraphics[width=1\columnwidth]{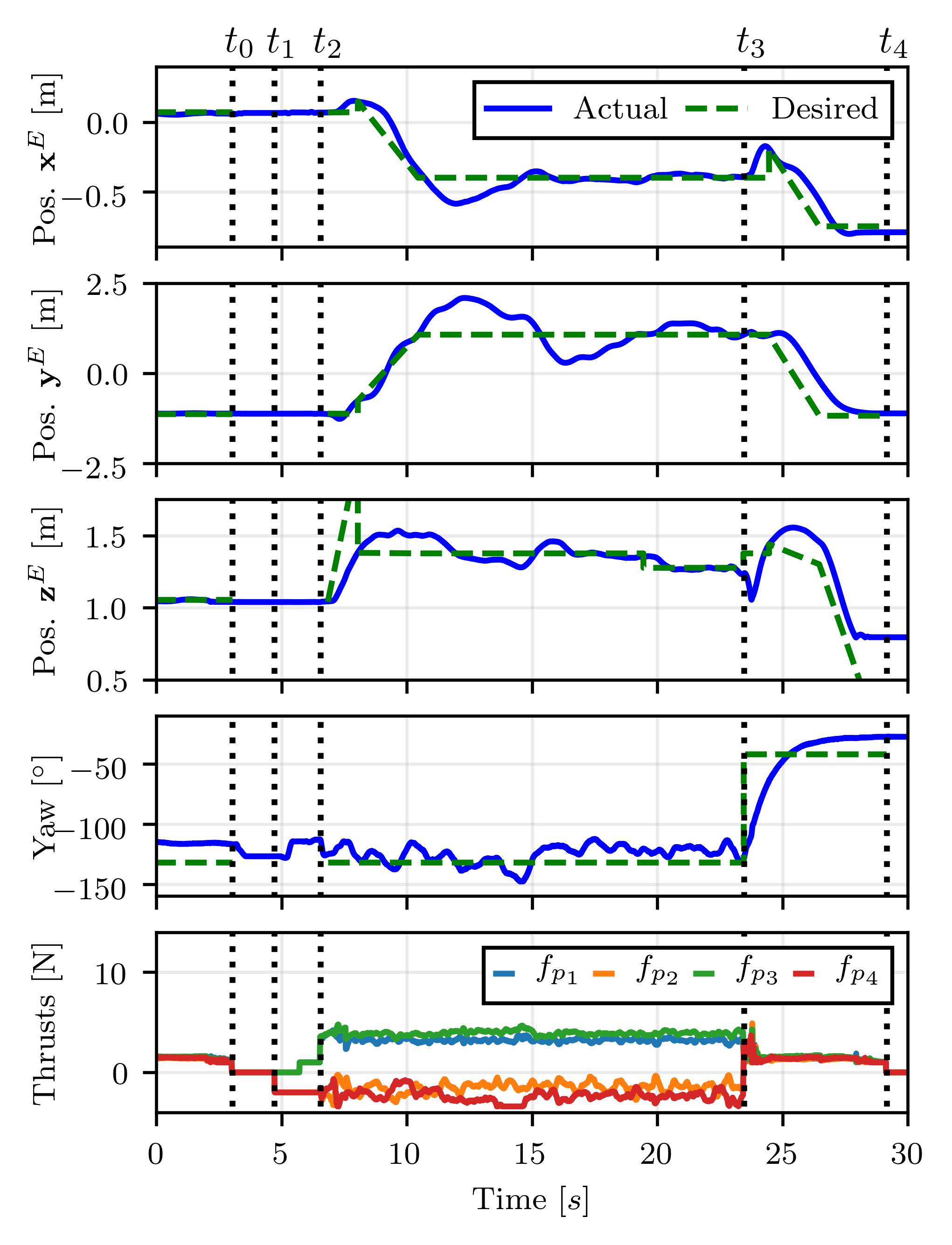}
	\centering
	\caption{Trajectory of the vehicle while performing a grasping task. The vehicle lands on the box at time $t_0$, and starts to grab the box at time $t_1$. At time $t_2$ the vehicle begins to lift the box, and flies to the desired drop-off location in the two-arms-folded configuration. Next, at time $t_3$, the vehicle transitions back to the unfolded configuration, dropping the box. Finally, the vehicle returns to its original position and lands at time $t_4$.}
	\label{fig:graspTraj}
\end{figure}

Figure~\ref{fig:graspTraj} shows the trajectory of the vehicle during the grasping maneuver.
Note that the vehicle experiences much larger position tracking errors in the two-arms-folded configuration compared to the tunnel traversal task shown in Figure~\ref{fig:tunnelTraj} (specifically in the $\y{}{E}$ direction).
Because the additional weight of the packages requires the two upward facing propellers to produce more thrust (and thus yaw torque), a larger yaw torque must be produced by the horizontally facing propellers in order to compensate.
This leads to occasional saturation of the thrust produced by the horizontal propellers, causing reduced tracking performance.
Additionally, because the additional mass of the package moves the center of mass of the vehicle closer to the thrust axes of the two horizontally facing propellers, the ability of these propellers to produce roll and pitch torques is reduced.
This effect is accounted for in the computation of \eqref{eq:twoArmsFoldedMixer} (which is used to synthesize the attitude controller), but nonetheless reduces the maximum control authority of the vehicle in roll and pitch compared to a vehicle without the package.
Finally, the size of the box both increases drag on the vehicle and results in significant aerodynamic interference between the box and the propellers, leading to reduced tracking accuracy.

Note that the grasping capabilities of the proposed vehicle are extremely limited compared to other designs, e.g., those described in Section \ref{sec:relWork}.
However, the proposed vehicle does not carry any sensors, actuators, or complex mechanisms beyond those used by a conventional quadcopter, and the limited grasping ability can thus be seen as a free (if very limited) capability.

%% file: conclusion.tex
\section{Conclusion} \label{sec:conclusion}

In this paper we have presented a novel quadcopter design that differs from a conventional quadcopter in the use of passive hinges which allow each of the four arms to rotate freely between unfolded and folded configurations.
The vehicle was designed to be nearly identical to a conventional quadcopter aside from the presence of the four passive hinges, resulting in the vehicle having a nearly identical power consumption as a conventional quadcopter when flying in the unfolded configuration.
Compared to a conventional quadcopter, additional constraints were placed on the control inputs of the vehicle such that the arms do not fold or unfold unexpectedly.
At hover, these constraints were shown to reduce the maximum yaw torque of our experimental vehicle by 36\%, but did not reduce the maximum thrust or roll and pitch torques.

A simple controller for the vehicle was proposed, which primarily differed from existing quadcopter controllers in the inclusion of the constraints used to prevent the arms from folding.
Additionally, a method for easily synthesizing controllers for the different configurations of the vehicle was presented and used to control the attitude of the vehicle in both the two- and four-arms-folded configurations.

The vehicle was designed such that it was capable of hovering in the two-arms-folded configuration.
Specifically, it was shown that the minimum angle of the arms relative to the central body is bounded by the characteristics of the propellers and the mass and size of the vehicle.
The agility of the experimental vehicle relative to a conventional quadcopter was quantified by analyzing the restrictiveness of the bounds used to prevent the arms from folding, showing a reduction in the maximum yaw torque of the vehicle.

Several experiments were conducted demonstrating the experimental vehicle performing different tasks.
We first compared the experimental vehicle to a conventional quadcopter, showing that the limits used to prevent the arms from folding have a negligible affect on trajectory tracking performance for a typical maneuver.
Next, we showed the experimental vehicle traversing a narrow, horizontal tunnel in the two-arms-folded configuration, highlighting how the vehicle transitions between configurations.
The vehicle was then shown traversing a vertical gap in the four-arms-folded configuration using projectile motion.
Finally, the vehicle was shown to be capable of perching on wires in the four-arms-folded configuration, and to be able to perform limited grasping tasks using the two-arms-folded configuration.

The proposed vehicle is thus able to track trajectories with effectively the same agility as a conventional quadcopter with the same mass and power properties, traverse narrow passageways, perch, and perform limited grasping.
For individual tasks, specialized designs (requiring, e.g., additional actuators and mass) may outperform the proposed vehicle, and for missions focused primarily on one task (e.g. grasping and carrying loads) these may be preferable.
However, for missions that primarily require flight in the unfolded configuration, the proposed vehicle has the ability to perform these additional tasks with very little trade-off in power consumption or agility compared to a conventional quadcopter.